\documentclass[11pt]{article}
\usepackage{jmlr2e}
\usepackage{abstract}
\usepackage{array}
\usepackage{bbding}
\usepackage{multirow}
\usepackage{adjustbox}
\usepackage{bbm}
\usepackage{float}
\usepackage{amsfonts}
\usepackage{graphicx}
\graphicspath{{graphics/}}
\usepackage{float}

\usepackage{tabularx,booktabs}
\usepackage{multirow}
\newcolumntype{Y}{>{\centering\arraybackslash}X}

\usepackage[ruled,vlined,algo2e]{algorithm2e}
\usepackage{algorithm}
\usepackage{algorithmic}
\usepackage{natbib}

\usepackage{lscape}
\usepackage{comment}
\usepackage{bm}
\usepackage[title]{appendix}
\usepackage{longtable}
\usepackage{mathtools}
\usepackage{dsfont}
\usepackage{fancyhdr}

\usepackage{stackengine}
\usepackage{textcomp}
\usepackage{stfloats}
\usepackage{verbatim}
\usepackage{xcolor}
\definecolor{darkblue}{rgb}{0.0,0.5,0.5}
\definecolor{blue}{rgb}{0.0,0.0,1}

\usepackage[colorlinks]{hyperref}
\hypersetup{colorlinks,breaklinks,linkcolor=blue,urlcolor=blue,anchorcolor=blue,citecolor=blue}
\usepackage{cleveref}
\usepackage{booktabs,caption}
\usepackage{multirow}
\usepackage{lscape}
\usepackage{epstopdf}
\usepackage{lineno}
\usepackage{microtype}

\usepackage{graphicx}
\usepackage{subcaption}

\usepackage{booktabs} % for professional tables
\usepackage{bbm}

\usepackage{amsfonts}
\usepackage{xurl}
\usepackage{stackengine}
\usepackage{tikz}
\usetikzlibrary{decorations.pathreplacing}
\usetikzlibrary{positioning,arrows.meta,quotes}
\usetikzlibrary{shapes,snakes}
\usetikzlibrary{bayesnet}
\tikzset{>=latex}
\tikzstyle{plate caption} = [caption, node distance=0, inner sep=0pt, below left=5pt and 0pt of #1.south]
\usepackage[normalem]{ulem}
\usepackage{multirow}

\usepackage{hyperref}
\usepackage{url}
\usepackage{enumitem}
\setlist{
  listparindent=\parindent,
  parsep=0pt,
}

\firstpageno{1}

\begin{document}

\title{Forecasting Sparse Movement Speed of Urban Road Networks with Nonstationary Temporal Matrix Factorization}

\author{\name Xinyu Chen\thanks{Corresponding author.} \email chenxy346@gmail.com \\
       \addr Polytechnique Montreal, Montreal, QC H3T 1J4, Canada
       \AND
       \name Chengyuan Zhang \email enzozcy@gmail.com \\
       \addr McGill University, Montreal, QC H3A 0C3, Canada
       \AND
       \name Xi-Le Zhao \email xlzhao122003@163.com \\
       \addr University of Electronic Science and Technology of China, Chengdu, Sichuan 611731, China
       \AND
       \name Nicolas Saunier \email nicolas.saunier@polymtl.ca \\
       \addr Polytechnique Montreal, Montreal, QC H3T 1J4, Canada
       \AND
       \name Lijun Sun \email lijun.sun@mcgill.ca \\
       \addr McGill University, Montreal, QC H3A 0C3, Canada}

\editor{}

\maketitle

\begin{abstract}
Movement speed data from urban road networks, computed from ridesharing vehicles or taxi trajectories, is often high-dimensional, sparse, and nonstationary (e.g., exhibiting seasonality). These characteristics pose challenges for developing scalable and efficient data-driven solutions for traffic flow estimation and forecasting using machine learning techniques. To address these challenges, we propose a Nonstationary Temporal Matrix Factorization (NoTMF) model that leverages matrix factorization to project high-dimensional and sparse movement speed data into low-dimensional latent spaces. This results in a concise formula with the multiplication between spatial and temporal factor matrices. To characterize the temporal correlations, NoTMF takes a latent equation on the seasonal differenced temporal factors using higher-order vector autoregression (VAR). This approach not only preserves the low-rank structure of sparse movement speed data but also maintains consistent temporal dynamics, including seasonality information. The learning process for NoTMF involves optimizing the spatial and temporal factor matrices along with a collection of VAR coefficient matrices. To solve this efficiently, we introduce an alternating minimization framework, which tackles a challenging procedure of estimating the temporal factor matrix using conjugate gradient method, as the subproblem involves both partially observed matrix factorization and seasonal differenced VAR. To evaluate the forecasting performance of NoTMF, we conduct extensive experiments on Uber movement speed datasets, which are estimated from ridesharing vehicle trajectories. These datasets contain a large proportion of missing values due to insufficient ridesharing vehicles on the urban road network. Despite the presence of missing data, NoTMF demonstrates superior forecasting accuracy and effectiveness compared to baseline models. Moreover, as the seasonality of movement speed data is of great concern, the experiment results highlight the significance of addressing the nonstationarity of movement speed data.

\end{abstract}

\begin{keywords}
Urban transportation network, ridesharing vehicles, floating car data, movement speed data, traffic flow forecasting, machine learning, matrix factorization, vector autoregression
\end{keywords}

\section{Introduction}

In the field of urban transportation networks, the rise of crowdsourced data, collected through fixed sensing detectors (e.g., loop detectors, radar detectors, and cameras) and mobile sensors (e.g., floating cars equipped with the Global Positioning System (GPS) devices), has revolutionized the monitoring of urban traffic states such as movement speeds \citep{treiber2013traffic, zheng2015trajectory}. Mobile sensor data, including human mobility trajectories, has gained great attention over the past decades due to the widespread utilization of mobile devices. These data provide unprecedented insights into analyzing human mobility behaviors (e.g., movement from place A to place B with a sequence of positions forming a trajectory) and enable the computation of movement speeds across urban road networks. For instance, trajectory data from ridesharing vehicles not only reveal the traffic states of urban road networks but also serve as pivotal components in city-wide traffic state estimation tasks. However, urban movement speed data derived from ridesharing vehicles are inherently sparse, capturing only a small fraction of the total vehicular trajectories and leading to the issues of missing values. 
% thus presenting considerable methodological challenges for accurate movement speed forecasting amidst data corruption and missing values.
% a limited number of entries contributed from a small percentage of vehicles of the whole traffic, namely, insufficient sampling. 
Thus, these imperfect data pose unprecedented methodological and practical challenges for accurate movement speed estimation and forecasting.

Using machine learning algorithms to estimate urban traffic flow from ridesharing trajectories is technically feasible but complex. The main challenge lies in leveraging partial observed movement speeds to accurately infer unobserved speeds. This challenge is central to the data-driven task of traffic flow imputation, which is critical for effective movement speed estimation \citep{chen2022bayesian}. An analysis of the Uber movement speed data\footnote{Available till October 2023 at \url{https://movement.uber.com/}.} highlights significance differences in data completeness between peak and off-peak hours, with higher missing rates of movement speeds during off-peak hours. Additionally, the data reveals daily seasonal patterns in missing values. These findings align with our understanding of urban traffic dynamics and emphasize the need to address data biases to improve the accuracy of urban traffic forecasting.

This study aims to advance the forecasting of sparse movement speeds within urban transportation networks by relying on several key assumptions: (i) Partially observed movement speed data contains informative low-dimensional spatial and temporal patterns; (ii) Movement speed data exhibits strong temporal dynamics and trends, which can be effectively modeled using linear time-invariant systems, such as vector autoregression (VAR); (iii) Addressing the nonstationarity of movement speed data through seasonal differencing, either directly in the data space or indirectly in latent spaces, is crucial for mitigating biases inherent in urban movement speed data. It is worth noting that matrix factorization is invariant to the permutation of rows and columns, which complicates the efforts to mitigate data bias. In the proposed NoTMF model, using seasonal differenced VAR is meaningful for characterizing traffic flow dynamics, such as global and local time series trends in the latent space, mitigating data biases (e.g., columns of data with varying missing rates, as shown in Figure~\ref{missing_rate_stat}), and reinforcing matrix factorization for more accurate imputation and prediction.

To summarize, the contributions of this study are threefold:
\begin{itemize}
\item We propose the NoTMF model, which integrates higher-order VAR processes with differencing operations into the classical low-rank matrix factorization framework. This approach enhances the model's ability to capture low-dimensional temporal dynamics within high-dimensional and sparse spatiotemporal datasets.
\item We design an alternating minimization algorithm for efficiently solving the optimization problem in the NoTMF model. The subproblem involving the low-dimensional temporal factor matrix---a generalized Sylvester equation that incorporates both partially observed matrix factorization and seasonal differenced VAR---is solved using the conjugate gradient method, which approximates the least squares solution of linear equations with minimal iterations.
\item We validate the NoTMF model's ability to forecast sparse movement speeds using two real-world datasets: the Uber movement speed data from New York City (NYC) and Seattle, USA, both of which encompass large numbers of road segments and exhibit strong seasonality. The experiments demonstrate the NoTMF model's superior performance in handling sparse data compared to existing state-of-the-art forecasting methods. In the meantime, the results highlight the importance of nonstationarity modeling with seasonal differenced VAR.
\end{itemize}

As we aim to formulate the sparse movement speed forecasting problem using data-driven machine learning methods, Table~\ref{notation} provides a summary of basic symbols and notations used in this study. Here, $\mathbb{R}$ represents the set of real numbers, while $\mathbb{Z}^+$ denotes the set of positive integers.

\begin{table*}[h!]
\centering
\caption{Summary of the basic notation.}
\label{notation}
\begin{tabular}{l|l} 
\toprule
Notation & Description \\ 
\midrule
$x\in\mathbb{R}$ & Scalar \\
$\boldsymbol{x}\in\mathbb{R}^{n}$ & Vector of length $n$ \\
$\boldsymbol{X}\in\mathbb{R}^{m\times n}$ & Matrix of size $m\times n$ \\
$\frac{\partial f}{\partial\boldsymbol{x}}$, $\frac{\partial f}{\partial\boldsymbol{X}}$ & Partial derivatives of $f$ with respect to vector $\boldsymbol{x}$ and matrix $\boldsymbol{X}$, respectively \\
$\delta\in\mathbb{Z}^{+}$ & Forecasting time horizon \\
$R\in\mathbb{Z}^{+}$ & Rank of matrix factorization \\
$[i]$ & Positive integer set $\{1,2,\ldots,i\},\,i\in\mathbb{Z}^{+}$ \\
$[0,i]$ & Nonnegative integer set $\{0,1,2,\ldots,i\},\,i\in\mathbb{Z}^{+}$ \\
$[i,j]$ & Positive integer set $\{i,i+1,\ldots,j\},\,i,j\in\mathbb{Z}^{+},\,i<j$ \\
$\|\cdot\|_2$ & $\ell_2$-norm of vector \\
$\|\cdot\|_F$ & Frobenius norm of matrix \\
$\otimes$ & Kronecker product, see Definition~\ref{kron_def} \\
$\operatorname{vec}(\cdot)$ & Vectorization operator \\
\bottomrule
\end{tabular}
\end{table*}

\section{Literature Review}

\subsection{Opportunities and Challenges of Using Ridesharing Vehicle Data in Traffic State Estimation}

In urban transportation networks, the utilization of fixed sensing detectors such as loop detectors, radar detectors, and cameras has been pivotal in monitoring traffic flow. These detectors are strategically placed to capture traffic data, collecting key parameters such as traffic density, flow, and speed. The effectiveness of these sensors in providing real-time traffic information has been well-documented in several studies \citep{jain2019review, guerrero2018sensor}. However, the advent of mobile sensor technology, including smartphones and vehicular sensors, has shifted the source of traffic data, offering a more dynamic and reliable approach to data collection \citep{janecek2015cellular}. 

Ridesharing vehicle data in the form of trajectories has been increasingly utilized to reveal traffic states across cities \citep{zheng2018traffic}, providing valuable insights into human mobility patterns. These patterns are not only help in understanding the movement from one position to another but also extend to computing movement speeds within urban road networks. Despite its potential, the reliance on mobile sensor data presents challenges, particularly concerning data sparsity and insufficient sampling. Ridesharing vehicles represent only a small fraction of the total traffic, resulting in datasets that are far from comprehensive. This sparsity is compounded by issues of data corruption and missing values, making it difficult to obtain accurate and reliable traffic states. As a result, there has been growing interest in developing methodologies capable of forecasting movement speeds despite these limitations. Substantial progress has been made in mitigating the effects of imperfect and sparse data, aiming to improve the accuracy and reliability of traffic state estimations derived from ridesharing vehicle data \citep{chen2022bayesian}. %When developing domain-specific machine learning algorithms, at least in the task of forecasting sparse urban movement speeds, one is also required to reduce data biases that would possibly mislead the loss function of machine learning models.

\subsection{Machine Learning Methods}

Matrix and tensor factorization techniques are widely used to explore traffic flow data. For example, \cite{yu2016temporal} establish a temporal regularized matrix factorization (TRMF) framework for forecasting high-dimensional and sparse traffic time series in latent spaces, while \cite{chen2022bayesian} develop a Bayesian temporal factorization for forecasting matrix and tensor time series data in the presence of missing values. The classical time series forecasting problem has been widely studied with a variety of statistics and machine learning frameworks over the past few decades. In this study, we aim to leverage matrix factorization and VAR to address the traffic flow forecasting problem with high-dimensional and sparse movement data.

In practice, on small-scale traffic time series datasets, VAR is a classical and efficient solution for traffic flow forecasting. However, when dealing with high-dimensional data, the traditional VAR model severely suffers from the over-parameterization issue \citep{velu1986reduced, velu1998multivariate, basu2015regularized, wang2021high, cheng2022real}, where the number of parameters vastly exceeds the number of observed data samples. In such cases, incorporating prior knowledge and assumptions about the parameter space, such as low-rank property and sparsity of the coefficient matrices, becomes essential for seeking an efficient learning process. For example, in multivariate reduced-rank regression \citep{velu1986reduced, velu1998multivariate}, the coefficient matrices are assumed to be low-rank, with variants of this method including high-dimensional VAR that use matrix factorization (e.g., \citep{carriero2016structural, koop2019bayesian}) and tensor factorization (e.g., \citep{wang2021high}). Imposing sparsity on the coefficients is another potential solution, such as using $\ell_1$-norm regularization \citep{basu2015regularized, han2015direct} or combining nuclear norm and sparsity-induced norms \citep{basu2019low}.

On incomplete time series data, a central challenge is to effectively learn temporal dynamics from partially observed data. A naive solution is to first impute the missing data and then make predictions on the estimated data points (e.g., \citep{che2018recurrent}). However, such methods may produce substantial estimation biases due to the separation of imputation and forecasting, not to mention that making accurate imputation itself is not a trivial task \citep{chen2022bayesian}. For high-dimensional datasets with missing values, the low-rank assumption has been demonstrated to be very effective for imputation \citep{yu2016temporal, chen2022bayesian}. In the literature, matrix and tensor factorization methods have been used to model high-dimensional and incomplete time series \citep{xiong2010temporal, yu2016temporal, takeuchi2017autoregressive, chen2022bayesian}. The underlying assumption in these models is that the incomplete high-dimensional dataset can be well characterized by a few latent factors---analogous to the state space representation \citep{stock2016dynamic} in dynamic factor models---which evolve over time following VAR processes. However, modeling the incomplete data in a matrix factorization framework is more computationally efficient due to the simplified isotropic covariance structure on errors \citep{yu2016temporal}.

In addition to the high-dimensionality and sparsity issues, nonstationarity is also a unique feature of real-world time series data. Recall that the properties of stationary time series do not depend on time. Most existing literature simply assumes that the data and the underlying latent temporal factors are stationary when applying VAR for modeling temporal dynamics \citep{yu2016temporal, takeuchi2017autoregressive, gultekin2018online, chen2022bayesian}. Only a few matrix/tensor factorization-based time series models address the nonstationarity issue by modeling the periodicity \citep{dunlavy2011temporal, yu2016temporal, kawabata2021ssmf}. As mentioned by \cite{kawabata2021ssmf}, to overcome the limitation of matrix factorization for seasonal modeling, one classical data augmentation approach for temporal modeling with periodic patterns is converting matrix factorization into tensor factorization according to season information. In the temporal regularized matrix factorization \citep{yu2016temporal}, seasonality is considered by using a well-designed lag set for the autoregressive model. Shifting seasonal matrix factorization proposed by \cite{kawabata2021ssmf} can learn multiple seasonal patterns/regimes from multi-viewed data, i.e., matrix-variate time series. These studies demonstrate the effectiveness of matrix and tensor factorization techniques for time series data with diverse periodic patterns. Nevertheless, to the best of our knowledge, introducing proper differencing operations---a simple yet effective solution to address the nonstationarity issue---has been overlooked by existing studies.

\subsection{Research Gaps}

By definition, one can represent the movement speed data collected from $N$ road segments and $T$ consecutive time steps (e.g., with an hourly time resolution) as the matrix $\boldsymbol{Y}\in\mathbb{R}^{N\times T}$ whose movement speeds at time $t$ are written as $\boldsymbol{y}_t\in\mathbb{R}^N$. The goal of movement speed forecasting is to estimate the future movement speeds $\hat{\boldsymbol{Y}}\in\mathbb{R}^{N\times \delta}$ for the next $\delta$ time steps ahead. A classical and widely used approach for multivariate time series forecasting is the VAR process \citep{lutkepohl2013introduction}:
\begin{equation}\label{var_formula}
\boldsymbol{y}_t=\sum_{k\in[d]}\boldsymbol{A}_k\boldsymbol{y}_{t-k}+\boldsymbol{\epsilon}_t,\,\forall t\in[d+1,T],
\end{equation}
where $d\in\mathbb{Z}^{+}$ represents the order of VAR, $\boldsymbol{A}_k\in\mathbb{R}^{N\times N},\, k\in[d]$ are the coefficient matrices, and $\boldsymbol{\epsilon}_t\in\mathbb{R}^N$ is the zero-mean Gaussian noise vector. With this formula, the coefficient matrices can capture coevolution movement patterns over different road segments. Typically, the standard VAR is well-suited to ``short-fat" data with $N\ll T$. However, the high-dimensionality has posed methodological challenges to the implementation of VAR due to the quadratically growing parameter space \citep{verleysen2005curse, basu2019low, wang2021high}. On the ``tall-skinny" data with $N\gg T$, it gives rise to the over-parameterization issue \citep{velu1986reduced, velu1998multivariate, basu2015regularized} with $dN^2$ parameters in \eqref{var_formula}, vastly exceeding $NT$ observations. Regarding the sparsity of urban movement speed data, it prevents us from using VAR given that the model usually requires complete observations for parameter estimation. 

To simultaneously address the high-dimensionality and sparsity issues, recent advances have integrated time series autoregression into the low-rank matrix/tensor factorization models \citep{xiong2010temporal, yu2016temporal, gultekin2018online, chen2022bayesian}, constructing a unified framework such as TMF. The TMF assumes that the movement speed data is dominated by a few ``important'' patterns (i.e., characterized by low-dimensional latent factors) and the temporal factor matrix captures the coevolution patterns of $N$ movement speed time series. In the literature, substantial progress has been made in verifying the effectiveness of TMF on many real-world time series data in the presence of missing values \citep{yu2016temporal, gultekin2018online, chen2022bayesian}. Nevertheless, some methodological challenges remain in learning the temporal dependencies. It is worth noting that the default TMF does not impose any constraints on the stationarity of latent temporal factors. This becomes problematic when modeling movement speed data with strong trends and seasonality. In this case, the estimated coefficient matrices are no longer effective because we would expect the underlying dynamics of temporal factors to be time-dependent and thus nonstationary. If the movement speed data is fully observed, a simple solution is to first perform certain differencing operations such as first-order differencing and seasonal differencing, and then use the differenced data for parameter estimation. However, this simple solution becomes inaccessible when the data is sparse. To this end, we intend to formulate the movement speed forecasting problem with NoTMF and present an efficient solution algorithm with alternating minimization and conjugate gradient methods.

\section{Nonstationary Temporal Matrix Factorization}

\subsection{Preliminaries: Matrix Factorization}

\begin{figure}[h!]
\centering
\includegraphics[scale=0.9]{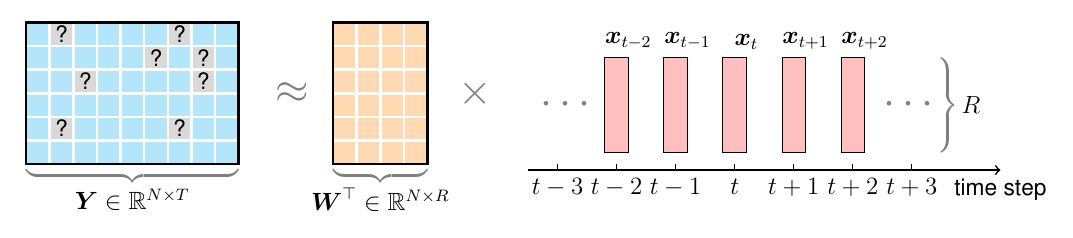}
\caption{Illustration of TMF on sparse movement speed data where the symbols ``?" represent the unobserved values in the data. TMF characterizes spatiotemporal patterns of the data $\boldsymbol{Y}\in\mathbb{R}^{N\times T}$ as a spatial factor matrix $\boldsymbol{W}\in\mathbb{R}^{R\times N}$ and a temporal factor matrix $\boldsymbol{X}\in\mathbb{R}^{R\times N}$ in which the temporal factor matrix is indeed a multivariate time series.}
\label{tmf}
\end{figure}

The movement speed data in urban road networks typically demonstrates certain low-dimensional spatial and temporal patterns. For instance, as shown in Figure~\ref{tmf}, the sparse movement speed data $\boldsymbol{Y}\in\mathbb{R}^{N\times T}$ can be approximated by the multiplication between a spatial factor matrix $\boldsymbol{W}\in\mathbb{R}^{R\times N}$ and a temporal factor matrix $\boldsymbol{X}\in\mathbb{R}^{R\times T}$, i.e., $\boldsymbol{Y}\approx\boldsymbol{W}^\top\boldsymbol{X}$ with a prescribed rank $R\in\mathbb{Z}^{+}<\min\{N, T\}$. In practice, the observed data $\boldsymbol{Y}$ is usually incomplete, which contains some missing/unobserved values in the data. Here we define $\mathcal{P}_{\Omega}(\cdot)$ as the orthogonal projection supported on the observed index set $\Omega$ to indicate the observed entries. For instance, on the data matrix $\boldsymbol{Y}$ with observed index set $\Omega$, we have the following definition for the orthogonal projection:
\begin{equation}
[\mathcal{P}_{\Omega}(\boldsymbol{Y})]_{n,t}=\begin{cases}
y_{n,t}, & \text{if $(n,t)\in\Omega$,} \\
0, & \text{otherwise},
\end{cases}
\end{equation}
where $n\in[N]$ and $t\in[T]$. Notably, $[\cdot]_{n,t}$ denotes the $(n,t)$-th entry of a matrix. The entries of $\boldsymbol{Y}$ filled with zeros imply the unobserved values, i.e., $(n,t)\notin\Omega$. Such a definition allows one to formulate the matrix factorization on the partially observed movement speed data.

As a consequence, the matrix factorization can be formulated as $\mathcal{P}_{\Omega}(\boldsymbol{Y})\approx\mathcal{P}_{\Omega}(\boldsymbol{W}^\top\boldsymbol{X})$ or element-wise $y_{n,t}\approx\boldsymbol{w}_n^\top\boldsymbol{x}_t,\,\forall n\in[N],\, t\in[T]$ in which $\boldsymbol{w}_n,\boldsymbol{x}_t\in\mathbb{R}^R$ are the $n$-th and the $t$-th columns of the factor matrices $\boldsymbol{W}$ and $\boldsymbol{X}$, respectively. To achieve such approximation, one  key problem is specifying the latent factor matrices, so that the partially specified data matrix $\boldsymbol{Y}$ matches $\boldsymbol{W}^\top\boldsymbol{X}$ as closely as possible \citep{koren2009matrix}. Formally, the optimization problem of matrix factorization is given by
\begin{equation}
\min_{\boldsymbol{W},\boldsymbol{X}}~\frac{1}{2}\bigl\|\mathcal{P}_{\Omega}(\boldsymbol{Y}-\boldsymbol{W}^\top\boldsymbol{X})\bigr\|_F^2+\frac{\rho}{2}(\|\boldsymbol{W}\|_F^2+\|\boldsymbol{X}\|_F^2),
\end{equation}
or on the observed entries of data matrix $\boldsymbol{Y}$, we have
\begin{equation}
\min_{\{\boldsymbol{w}_n\}_{n\in[N]},\{\boldsymbol{x}_{t}\}_{t\in[T]}}~\frac{1}{2}\sum_{(n,t)\in\Omega}(y_{n,t}-\boldsymbol{w}_{n}^\top\boldsymbol{x}_t)^2+\frac{\rho}{2}\Bigl(\sum_{n\in[N]}\|\boldsymbol{w}_{n}\|_2^2+\sum_{t\in[T]}\|\boldsymbol{x}_t\|_2^2\Bigr),
\end{equation}
with an objective function defined by the error of matrix factorization on the set $\Omega$ and the regularization terms (weighted by the hyperparameter $\rho$) corresponding to factor matrices $\boldsymbol{W}$ and $\boldsymbol{X}$. Alternatively, the column vector $\boldsymbol{y}_{t}\in\mathbb{R}^{N},\,\forall t\in[T]$ of the data $\boldsymbol{Y}$ can be approximated by $\boldsymbol{W}^\top\boldsymbol{x}_{t}$. To characterize the temporal dynamics of urban movement speed, we next introduce the NoTMF model which integrates the seasonal differenced VAR into the matrix factorization framework. By doing so, NoTMF can build complicated temporal correlations when forecasting high-dimensional and sparse movement speeds of urban road networks.

\subsection{Optimization Problem of NoTMF}

TMF that integrated temporal modeling techniques such as univariate autoregression and VAR has become extremely useful for multivariate time series forecasting in the presence of missing values \citep{yu2016temporal, gultekin2018online, chen2022bayesian}. The essential idea of the TMF models is that low-rank matrix factorization can discover low-dimensional temporal patterns from partially observed time series, while autoregression can reinforce the temporal correlations in the modeling process and capture time-evolving coefficients. To this end, we first characterize time series correlations and patterns from sparse time series data and then make efficient predictions for future data points. At the same time, to address the real-world sparse prediction problem, the design of TMF stems from some classical machine learning frameworks such as online learning and dictionary learning.

TMF is well-suited to handling the following emerging issues in real-world movement speed data: 1) High-dimensionality: Urban road networks such as transportation systems in NYC usually have thousands of road segments, i.e., large $N$; 2) Sparsity: Road-level movement speed data are often sparse, which means that only a small portion of data are observed due to the limited penetration of float cars (e.g., taxi and ridesharing vehicle). On Uber movement speed datasets, we only have access to a small portion of movement speed values even in an hourly time resolution due to insufficient sampling and the limited penetration of ridesharing vehicles. %In what follows, we let $\Omega$ be the observed index set of the movement speed data $\boldsymbol{Y}\in\mathbb{R}^{N\times T}$.

Recall that an essential assumption of VAR is that the modeled time series is stationary. However, the periodicity/seasonality in real-world movement speed data $\boldsymbol{Y}$ will be characterized by the latent temporal factor matrix $\boldsymbol{X}$, and such nonstationarity property contradicts the stationary assumption in VAR. Furthermore, the estimated coefficients in the VAR process would become sub-optimal and even produce biased estimations for the movement speed. To address this issue, we expect the temporal dynamics to be time-dependent and introduce seasonal differencing to stationarize traffic time series dynamics in latent spaces. For any sparse movement speed data $\boldsymbol{Y}\in\mathbb{R}^{N\times T}$ collected from $N$ road segments and $T$ (hourly) time steps, we formulate the NoTMF model as the following optimization problem:
\begin{equation}\label{optimization_problem_v1}
\begin{aligned}
\min_{\boldsymbol{W},\,\boldsymbol{X},\,\{\boldsymbol{A}_k\}_{k\in[d]}}\frac{1}{2}\bigl\|\mathcal{P}_{\Omega}(\boldsymbol{Y}-\boldsymbol{W}^\top\boldsymbol{X})\bigr\|_F^2 +\frac{\gamma}{2}\sum_{t\in[d+m+1,T]}\Bigl\|\dot{\boldsymbol{x}}_t^{m}-\sum_{k\in[d]}\boldsymbol{A}_{k}\dot{\boldsymbol{x}}_{t-k}^{m}\Bigr\|_2^2 +\frac{\rho}{2}(\|\boldsymbol{W}\|_F^2+\|\boldsymbol{X}\|_F^2),\\
\end{aligned}
\end{equation}
where the season is set as $m\in\mathbb{Z}^{+}$ and we let
\begin{equation}
\dot{\boldsymbol{x}}_{t}^{m}\triangleq\boldsymbol{x}_{t}-\boldsymbol{x}_{t-m},\,\forall t\in[m+1,T],
\end{equation}
be the result of the season-$m$ differencing on the columns of the temporal factor matrix $\boldsymbol{X}$. On the seasonal different temporal factors, $\boldsymbol{A}_{k}\in\mathbb{R}^{R\times R},\, k\in[d]$ represent the coefficient matrices of $d$th-order VAR for characterizing the temporal dynamics of $\boldsymbol{X}$. The errors of seasonal differenced VAR are formulated as a regularization term, leading to the design of temporal loss. Hyperparameters $\{\gamma,\rho\}$ are the weight parameters for the regularization terms.

In practice, the objective function in \eqref{optimization_problem_v1} has three components, including the loss function of matrix factorization, the temporal loss of seasonal differenced VAR, and the regularization terms. The matrix factorization can help recover missing values in the matrix $\boldsymbol{Y}$ through $\boldsymbol{W}^\top\boldsymbol{X}$, but the data biases (e.g., the columns with different amounts of observed entries) possibly mislead the loss function behaviors. Thus, the temporal loss is of great significance for mitigating data biases of urban movement speeds. In theory, our NoTMF takes the form of linear time-invariant systems and has an observation equation and a latent space equation, namely,
\begin{equation}\label{obs_vs_latent_eqs}
\begin{cases}
\mathcal{P}_{\Omega_t}(\boldsymbol{y}_t)=\mathcal{P}_{\Omega_t}(\boldsymbol{W}^\top\boldsymbol{x}_t)+\boldsymbol{\epsilon}_t, & \text{(Observation equation)} \\
\displaystyle\dot{\boldsymbol{x}}_t^{m}=\sum_{k\in[d]}\boldsymbol{A}_k\dot{\boldsymbol{x}}_{t-k}^{m}+\boldsymbol{\eta}_t, & \text{(Latent space equation)}
\end{cases}
\end{equation}
where $\Omega_{t}$ is the observed index set of the high-dimensional movement speeds $\boldsymbol{y}_t\in\mathbb{R}^N$ at time $t$. The vectors $\boldsymbol{\epsilon}_t\in\mathbb{R}^{N}$ and $\boldsymbol{\eta}_t\in\mathbb{R}^{R}$ are zero-mean Gaussian noises. As can be seen, the observation equation describes the relation between the partially observed variables $\boldsymbol{y}_t,\,\forall t\in[T]$ and spatial/temporal factors with matrix factorization, while the latent space equation reflects the dynamics of the unobserved factor variables $\boldsymbol{x}_t,\,\forall t\in[T]$ with VAR. A phenomenon related to \eqref{obs_vs_latent_eqs} is the existence of ``representative" temporal dynamics of urban traffic flow which comes from thousands of road segments. Despite the sparse learning process in NoTMF, the fundamental idea of NoTMF can be easily connected with some classical methods such as dynamic factor model in macroeconomic modeling \citep{forni2000generalized, stock2016dynamic}, state space model in system identification and control \citep{hangos2006analysis}, and recurrent neural networks in deep learning \citep{lecun2015deep, prince2023understanding}.

\subsection{Matrix Representation: Temporal Operator Matrices}

As mentioned above, the optimization problem of NoTMF has a matrix factorization component in the form of matrices and a VAR component in the form of vectors. To get a unified solution to the optimization problem, we rewrite the vector-form temporal loss in the form of matrices by utilizing the temporal operator matrices as described in Definition~\ref{temporal_operator_def}.

\begin{definition}[Temporal Operator Matrices]\label{temporal_operator_def}
According to the expression of $d$th-order VAR on the latent temporal factors $\{\boldsymbol{x}_{t}\}_{t\in[T]}$, the temporal operator matrices $\{\boldsymbol{\Psi}_{k}\}_{k\in[0,d]}$ associated with the season $m\in\mathbb{Z}^{+}$ are defined as follows,
\begin{equation}\label{Psi_k_def}
\begin{aligned}
\boldsymbol{\Psi}_k\triangleq &\begin{bmatrix}
\boldsymbol{0}_{(T-d-m)\times (d-k)}\ \  & -\boldsymbol{I}_{T-d-m} \ \ & \boldsymbol{0}_{(T-d-m)\times (k+m)}
\end{bmatrix} \\
&+\begin{bmatrix}
\boldsymbol{0}_{(T-d-m)\times (d+m-k)}\ \  & \boldsymbol{I}_{T-d-m}\ \  & \boldsymbol{0}_{(T-d-m)\times k}
\end{bmatrix} \\
&\in\mathbb{R}^{(T-d-m)\times T},\,\forall k\in[0,d],
\end{aligned}
\end{equation}
where $\boldsymbol{0}_{m\times n}$ denotes the $m$-by-$n$ matrix of zeros, and $\boldsymbol{I}_{m}$ denotes the $m$-by-$m$ identity matrix. In \eqref{Psi_k_def}, the temporal operator matrix $\boldsymbol{\Psi}_k$ is an augmented matrix, i.e., the combination of two matrices. In particular, the first matrix has zeros as the entries in the first $d-k$ columns and the last $k+m$ columns, while the second matrix has zeros as the entries in the first $d+m-k$ columns and the last $k$ columns.
\end{definition}

\begin{remark}
Since these temporal operator matrices $\{\boldsymbol{\Psi}_k\}_{k\in[0,d]}$ are a sequence of sparse matrices and only have $2(T-d-m)$ nonnegative entries, utilizing such sparse structure allows one to reduce the memory consumption and develop efficient algorithms in the case of a very long time series, i.e., large $T$.
\end{remark}

As a result, one can use the multiplication between the temporal factor matrix $\boldsymbol{X}$ and the temporal operator matrices to reformulate the temporal loss. First, the season-$m$ differenced temporal factors can be written in the form of matrices:
\begin{equation}
\begin{aligned}
\boldsymbol{X}\boldsymbol{\Psi}_0^\top&=\begin{bmatrix}
\mid & & \mid \\
\boldsymbol{x}_{d+m+1}-\boldsymbol{x}_{d+1} & \cdots & \boldsymbol{x}_{T}-\boldsymbol{x}_{T-m} \\
\mid & & \mid \\
\end{bmatrix}, \\
\boldsymbol{X}\boldsymbol{\Psi}_1^\top&=\begin{bmatrix}
\mid & & \mid \\
\boldsymbol{x}_{d+m}-\boldsymbol{x}_{d} & \cdots & \boldsymbol{x}_{T-1}-\boldsymbol{x}_{T-m-1} \\
\mid & & \mid \\
\end{bmatrix}, \\
&~\vdots \\
\boldsymbol{X}\boldsymbol{\Psi}_d^\top&=\begin{bmatrix}
\mid & & \mid \\
\boldsymbol{x}_{m+1}-\boldsymbol{x}_{1} & \cdots & \boldsymbol{x}_{T-d}-\boldsymbol{x}_{T-m-d} \\
\mid & & \mid \\
\end{bmatrix}, \\
\end{aligned}
\end{equation}
of size $R\times (T-d-m)$. Then, one can eliminate the sum over all $t\in[d+m+1, T]$ in the temporal loss and find an equivalent expression, namely,
\begin{equation}
\sum_{t\in[d+m+1,T]}\Bigl\|\dot{\boldsymbol{x}}_{t}^{m}-\sum_{k\in[d]}\boldsymbol{A}_k\dot{\boldsymbol{x}}_{t-k}^{m}\Bigr\|_{2}^{2}=\Bigl\|\boldsymbol{X}\boldsymbol{\Psi}_0^\top-\sum_{k\in[d]}\boldsymbol{A}_{k}\boldsymbol{X}\boldsymbol{\Psi}_k^\top\Bigr\|_F^2.
\end{equation}

By doing so, the matrix-form temporal loss can facilitate the matrix computations. Thus, the optimization problem of NoTMF in \eqref{optimization_problem_v1} now becomes
\begin{equation}\label{optimization_problem_v2}
\begin{aligned}
\min_{\boldsymbol{W},\,\boldsymbol{X},\,\{\boldsymbol{A}_k\}_{k\in[d]}}\frac{1}{2}\bigl\|\mathcal{P}_{\Omega}(\boldsymbol{Y}-\boldsymbol{W}^\top\boldsymbol{X})\bigr\|_F^2 +\frac{\gamma}{2}\Bigl\|\boldsymbol{X}\boldsymbol{\Psi}_0^\top-\sum_{k\in[d]}\boldsymbol{A}_{k}\boldsymbol{X}\boldsymbol{\Psi}_k^\top\Bigr\|_F^2+\frac{\rho}{2}(\|\boldsymbol{W}\|_F^2+\|\boldsymbol{X}\|_F^2), \\
\end{aligned}
\end{equation}
where the latent factor matrices $\{\boldsymbol{W},\boldsymbol{X}\}$ and the coefficient matrices $\{\boldsymbol{A}_k\}_{k\in[d]}$ are required to estimate. If one obtains both spatial factor matrix $\boldsymbol{W}$ and temporal factor matrix $\boldsymbol{X}$, then the unobserved entries in the data $\boldsymbol{Y}$ can be reconstructed accordingly. In particular, the VAR process can reinforce the temporal correlations among $\boldsymbol{x}_{t},\,\forall t\in[T]$ and capture temporal dynamics, while the matrix factorization without temporal modeling is invariant to the permutation of column vectors $\{\boldsymbol{x}_{t}\}_{t\in[T]}$ in the temporal factor matrix $\boldsymbol{X}$. We next introduce an alternating minimization framework to solve the optimization problem of NoTMF in \eqref{optimization_problem_v2}, in which we denote the objective function by $f$.

\subsection{Spatial Factor Matrix Estimation}

The spatial factor matrix $\boldsymbol{W}$ represents low-dimensional patterns of the sparse movement speed data $\boldsymbol{Y}$. In the alternating minimization framework, estimating the spatial factor matrix $\boldsymbol{W}$ requires one to fix the temporal factor matrix $\boldsymbol{X}$ and the coefficient matrices $\{\boldsymbol{A}_{k}\}_{k\in[d]}$ as known variables. First of all, we write down the partial derivative of $f$ with respect to $\boldsymbol{W}$ as follows,
\begin{equation}
\frac{\partial f}{\partial\boldsymbol{W}}=-\boldsymbol{X}\mathcal{P}_{\Omega}^\top(\boldsymbol{Y}-\boldsymbol{W}^\top\boldsymbol{X})+\rho\boldsymbol{W}.
\end{equation}

Let $\frac{\partial f}{\partial\boldsymbol{W}}=\boldsymbol{0}$, then we have the following generalized Sylvester equation:
\begin{equation}\label{matrix_equation_w}
\boldsymbol{X}\mathcal{P}_{\Omega}^\top(\boldsymbol{W}^\top\boldsymbol{X})+\rho\boldsymbol{W}=\boldsymbol{X}\mathcal{P}_{\Omega}^\top(\boldsymbol{Y}).
\end{equation}

In fact, \eqref{matrix_equation_w} is a generalized system of linear equations, and it is not difficult to obtain the closed-form solution. A natural solution could be pursued through the least squares of each column vector of $\boldsymbol{W}$ \citep{koren2009matrix}:
\begin{equation}\label{solution_to_w}
\boldsymbol{w}_{n}=\Bigl(\sum_{t:(n,t)\in\Omega}\boldsymbol{x}_t\boldsymbol{x}_t^\top+\rho\boldsymbol{I}_{R}\Bigr)^{-1}\Bigl(\sum_{t:(n,t)\in\Omega}\boldsymbol{x}_ty_{n,t}\Bigr),\,\forall n\in[N],
\end{equation}
where the notation $\displaystyle\sum_{t:(n,t)\in\Omega}$ implies the sum over all $t$ in the observed index set $\Omega$ with the fixed index $n$.

\subsection{Temporal Factor Matrix Estimation: Factorization Plus Autoregression}

The temporal factor matrix $\boldsymbol{X}$ represents low-dimensional dynamic patterns of the sparse movement speed data $\boldsymbol{Y}$. We assume a seasonal differencing on the nonstationary temporal factor matrix and assert that such a VAR process can reinforce the temporal modeling capability of matrix factorization on sparse movement speed data. However, the challenge would arise because VAR on the seasonal differenced temporal factor matrix complicates the optimization problem. With respect to the temporal factor matrix $\boldsymbol{X}$, the partial derivative of $f$ can be written as follows,
\begin{equation}
\frac{\partial f}{\partial\boldsymbol{X}}=-\boldsymbol{W}\mathcal{P}_{\Omega}(\boldsymbol{Y}-\boldsymbol{W}^\top\boldsymbol{X})+\gamma\sum_{k\in[0,d]}\boldsymbol{A}_k^\top\Bigl(\sum_{h\in[0,d]}\boldsymbol{A}_h\boldsymbol{X}\boldsymbol{\Psi}_h^\top\Bigr)\boldsymbol{\Psi}_k+\rho\boldsymbol{X},
\end{equation}
where $\boldsymbol{A}_0\triangleq-\boldsymbol{I}_R$ is introduced as a negative identity matrix.

Let $\frac{\partial f}{\partial\boldsymbol{X}}=\boldsymbol{0}$, then we have a generalized Sylvester equation with multiple terms:
\begin{equation}\label{matrix_equation_x}
\boldsymbol{W}\mathcal{P}_{\Omega}(\boldsymbol{W}^\top\boldsymbol{X})+\gamma\sum_{k\in[0,d]}\boldsymbol{A}_{k}^\top\Bigl(\sum_{h\in[0,d]}\boldsymbol{A}_{h}\boldsymbol{X}\boldsymbol{\Psi}_h^\top\Bigr)\boldsymbol{\Psi}_k+\rho\boldsymbol{X}=\boldsymbol{W}\mathcal{P}_{\Omega}(\boldsymbol{Y}),
\end{equation}
which involves $d^2+2$ terms in the left-hand side. Both orthogonal projection in matrix factorization and VAR on the temporal factor matrix complicate this matrix equation. If we convert the formula into a generalized system of linear equations, then finding the closed-form solution to \eqref{matrix_equation_x} would take $\mathcal{O}(R^3T^3)$ time, infeasible as the dimension $RT$ is large, see details as shown in Lemma~\ref{mixed_product_kron_lemma}.

\begin{lemma}
Suppose $\boldsymbol{A}\in\mathbb{R}^{m\times n}$, $\boldsymbol{X}\in\mathbb{R}^{n\times p}$, and $\boldsymbol{B}\in\mathbb{R}^{p\times q}$ be three matrices commensurate from multiplication in that order, then it always hold that
\begin{equation}
\operatorname{vec}(\boldsymbol{A}\boldsymbol{X}\boldsymbol{B})=(\boldsymbol{B}^\top\otimes\boldsymbol{A})\operatorname{vec}(\boldsymbol{X}),
\end{equation}
where $\operatorname{vec}(\cdot)$ denotes the vectorization operator, and $\otimes$ denotes the Kronecker product (see Definition~\ref{kron_def}). The aforementioned formula is also known as the mixed-product property of Kronecker product \citep{golubl2013matrix}.
\end{lemma}

\begin{definition}[Kronecker Product \citep{golubl2013matrix}]\label{kron_def}
For any matrices $\boldsymbol{A}\in\mathbb{R}^{m\times n}$ and $\boldsymbol{B}\in\mathbb{R}^{p\times q}$, the Kronecker product between $\boldsymbol{A}$ and $\boldsymbol{B}$ is given by
\begin{equation}
\boldsymbol{A}\otimes\boldsymbol{B}=\begin{bmatrix}
a_{11}\boldsymbol{B} & a_{12}\boldsymbol{B} & \cdots & a_{1n}\boldsymbol{B} \\
a_{21}\boldsymbol{B} & a_{22}\boldsymbol{B} & \cdots & a_{2n}\boldsymbol{B} \\
\vdots & \vdots & \ddots & \vdots \\
a_{m1}\boldsymbol{B} & a_{m2}\boldsymbol{B} & \cdots & a_{mn}\boldsymbol{B} \\
\end{bmatrix}\in\mathbb{R}^{(mp)\times (nq)},
\end{equation}
where the resulting matrix is of size $(mp)\times (nq)$ with $m\times n$ blocks, and each block is the multiplication between the matrix $\boldsymbol{B}$ and the entry of $\boldsymbol{A}$.
\end{definition}

\begin{lemma}\label{mixed_product_kron_lemma}
Given any partially observed matrix $\boldsymbol{Y}\in\mathbb{R}^{N\times T}$ with the observed index set $\Omega$, the closed-form solution to \eqref{matrix_equation_x} in the form of vectorized $\boldsymbol{X}$ is given by
\begin{equation}\label{solution_to_x}
\operatorname{vec}(\boldsymbol{X})=\Bigl(\boldsymbol{S}+\gamma\sum_{k\in[0,d]}\sum_{h\in[0,d]}(\boldsymbol{\Psi}_k\otimes\boldsymbol{A}_k)^\top(\boldsymbol{\Psi}_h\otimes\boldsymbol{A}_h)+\rho\boldsymbol{I}_{RT}\Bigr)^{-1}\operatorname{vec}(\boldsymbol{W}\mathcal{P}_{\Omega}(\boldsymbol{Y})),
\end{equation}
where the block matrix
\begin{equation*}
\boldsymbol{S}\triangleq\begin{bmatrix}
\boldsymbol{S}_1 & \boldsymbol{0} & \cdots & \boldsymbol{0} \\
\boldsymbol{0} & \boldsymbol{S}_2 & \cdots & \boldsymbol{0} \\
\vdots & \vdots & \ddots & \vdots \\
\boldsymbol{0} & \boldsymbol{0} & \cdots & \boldsymbol{S}_{T} \\
\end{bmatrix}\in\mathbb{R}^{(RT)\times (RT)},
\end{equation*}
has a sequence of blocks on the diagonal such that
\begin{equation*}
\boldsymbol{S}_{t}\triangleq\sum_{n:(n,t)\in\Omega}\boldsymbol{w}_n\boldsymbol{w}_n^\top\in\mathbb{R}^{R\times R},\,\forall t\in[T],
\end{equation*}
where the notation $\displaystyle\sum_{n:(n,t)\in\Omega}$ implies the sum over all $n$ in the observed index set $\Omega$ with the fixed index $t$.
\end{lemma}

\begin{remark}
If we let the first term of the left-hand side in \eqref{matrix_equation_x} as
\begin{equation}
\boldsymbol{C}\triangleq\boldsymbol{W}\mathcal{P}_{\Omega}(\boldsymbol{W}^\top\boldsymbol{X})\in\mathbb{R}^{R\times T},
\end{equation}
then the $t$-th column vector $\boldsymbol{c}_t\in\mathbb{R}^{R}$ of the matrix $\boldsymbol{C}$ is given by
\begin{equation}
\begin{aligned}
\boldsymbol{c}_{t}=\sum_{n:(n,t)\in\Omega}\boldsymbol{w}_n\boldsymbol{w}_n^\top\boldsymbol{x}_t=\boldsymbol{S}_{t}\boldsymbol{x}_t \quad
\Rightarrow\quad\operatorname{vec}(\boldsymbol{C})=\boldsymbol{S}\operatorname{vec}(\boldsymbol{X}).
\end{aligned}
\end{equation}

In the meantime, if we let the VAR components in \eqref{matrix_equation_x} be
\begin{equation}
\boldsymbol{E}\triangleq\sum_{k\in[0,d]}\boldsymbol{A}_{k}^\top\Bigl(\sum_{h\in[0,d]}\boldsymbol{A}_h\boldsymbol{X}\boldsymbol{\Psi}_h^\top\Bigr)\boldsymbol{\Psi}_k\in\mathbb{R}^{R\times T},
\end{equation}
then according to the mixed-product property of Kronecker product in Lemma~\ref{mixed_product_kron_lemma}, we have
\begin{equation}
\begin{aligned}
\operatorname{vec}(\boldsymbol{E})=&\sum_{k\in[0,d]}(\boldsymbol{\Psi}_k^\top\otimes\boldsymbol{A}_k^\top)\operatorname{vec}\Bigl(\sum_{h\in[0,d]}\boldsymbol{A}_h\boldsymbol{X}\boldsymbol{\Psi}_h^\top\Bigr) \\
=&\sum_{k\in[0,d]}\sum_{h\in[0,d]}(\boldsymbol{\Psi}_k\otimes\boldsymbol{A}_k)^\top(\boldsymbol{\Psi}_h\otimes\boldsymbol{A}_h)\operatorname{vec}(\boldsymbol{X}).
\end{aligned}
\end{equation}
As a consequence, \eqref{matrix_equation_x} is equivalent to 
\begin{equation}\label{matrix_equation_x_vector_form}
\Bigl(\boldsymbol{S}+\gamma\sum_{k\in[0,d]}\sum_{h\in[0,d]}(\boldsymbol{\Psi}_k\otimes\boldsymbol{A}_k)^\top(\boldsymbol{\Psi}_h\otimes\boldsymbol{A}_h)+\rho\boldsymbol{I}_{RT}\Bigr)\operatorname{vec}(\boldsymbol{X})=\operatorname{vec}(\boldsymbol{W}\mathcal{P}_{\Omega}(\boldsymbol{Y})),
\end{equation}
and thus leading to \eqref{solution_to_x} as claimed.
\end{remark}

However, the vectorized closed-form solution in \eqref{solution_to_x} is memory-consuming and computationally expensive for large problems. In large applications, as suggested by \cite{rao2015collaborative, chi2019nonconvex}, the alternating minimization with conjugate gradient for solving matrix factorization problems is both efficient and scalable. The conjugate gradient method allows one to search for the approximated solution to a system of linear equations with a relatively small number of iterations (e.g., 5 or 10). Therefore, to estimate the temporal factor matrix $\boldsymbol{X}$, we develop an efficient conjugate gradient routine. We first define the left-hand side of \eqref{matrix_equation_x_vector_form} as the function $\theta_{x}:\mathbb{R}^{R\times T}\to\mathbb{R}^{RT}$, whose formula is given by
\begin{equation}
\theta_{x}(\boldsymbol{X})\triangleq \Bigl(\boldsymbol{S}+\gamma\sum_{k\in[0,d]}\sum_{h\in[0,d]}(\boldsymbol{\Psi}_k\otimes\boldsymbol{A}_k)^\top(\boldsymbol{\Psi}_h\otimes\boldsymbol{A}_h)+\rho\boldsymbol{I}_{RT}\Bigr)\operatorname{vec}(\boldsymbol{X}),
\end{equation}
which is equivalent to
\begin{equation}
\theta_{x}(\boldsymbol{X})=\operatorname{vec}\Bigl(\boldsymbol{W}\mathcal{P}_{\Omega}(\boldsymbol{W}^\top\boldsymbol{X})+\gamma\sum_{k\in[0,d]}\boldsymbol{A}_{{k}}^\top\Bigl(\sum_{h\in[0,d]}\boldsymbol{A}_{h}\boldsymbol{X}\boldsymbol{\Psi}_h^\top\Bigr)\boldsymbol{\Psi}_k+\rho\boldsymbol{X}\Bigr).
\end{equation}

Accordingly, we let the right-hand side of \eqref{matrix_equation_x_vector_form} be
\begin{equation}
\psi_x\triangleq\operatorname{vec}(\boldsymbol{W}\mathcal{P}_{\Omega}(\boldsymbol{Y})).
\end{equation}

As mentioned above, \eqref{matrix_equation_x_vector_form} is in the form of a system of linear equations, and its system matrix
% \begin{equation*}
$\boldsymbol{S}+\gamma\sum_{k\in[0,d]}\sum_{h\in[0,d]}(\boldsymbol{\Psi}_k\otimes\boldsymbol{A}_k)^\top(\boldsymbol{\Psi}_h\otimes\boldsymbol{A}_h)+\rho\boldsymbol{I}_{RT}$
% \end{equation*}
is a symmetric positive-definite matrix. The conjugate gradient method is well-suited to such Kronecker-structured linear equations. Algorithm~\ref{conj_grad_to_X_algo} summarizes the estimation procedure for approximating the least squares solution to $\boldsymbol{X}$ in \eqref{solution_to_x}. Alternatively, we can use the conjugate gradient to solve \eqref{matrix_equation_w} when estimating the spatial factor matrix $\boldsymbol{W}$.

\begin{algorithm}
{%\fontsize{10pt}{10pt}\selectfont
\caption{Conjugate Gradient for Approximating $\boldsymbol{X}$}
\label{conj_grad_to_X_algo}
\begin{algorithmic}[1]
 \renewcommand{\algorithmicrequire}{\textbf{Input:}}
 \renewcommand{\algorithmicensure}{\textbf{Output:}}
 \REQUIRE Data matrix $\boldsymbol{Y}$ with the observed index set $\Omega$, spatial factor matrix $\boldsymbol{W}$, initialized temporal factor matrix $\boldsymbol{X}$, coefficient matrices $\{\boldsymbol{A}_k\}_{k\in[d]}$, and maximum iteration $\ell_{\max}$ (e.g., $\ell_{\max}=5$).
 \ENSURE Estimated temporal factor matrix $\boldsymbol{X}$.
 \STATE Initialize $\boldsymbol{x}_{0}$ by the vectorized $\boldsymbol{X}$.
 \STATE Compute residual vector $\boldsymbol{r}_{0}=\psi_x-\theta_x(\boldsymbol{X})$, and let $\boldsymbol{d}_0=\boldsymbol{r}_0$.
  \FOR {$\ell = 0$ to $\ell_{\max}-1$}
  \STATE Convert vector $\boldsymbol{d}_{\ell}$ into matrix $\boldsymbol{D}_{\ell}$.
  \STATE Compute $\alpha_{\ell}=\frac{\boldsymbol{r}_{\ell}^\top\boldsymbol{r}_{\ell}}{\boldsymbol{d}_{\ell}^\top\theta_x(\boldsymbol{D}_{\ell})}$.
  \STATE Update $\boldsymbol{x}_{\ell+1}=\boldsymbol{x}_{\ell}+\alpha_{\ell}\boldsymbol{d}_{\ell}$.
  \STATE Update $\boldsymbol{r}_{\ell+1}=\boldsymbol{r}_{\ell}-\alpha_{\ell}\theta_x(\boldsymbol{D}_{\ell})$.
  \STATE Compute $\beta_{\ell}=\frac{\boldsymbol{r}_{\ell+1}^\top\boldsymbol{r}_{\ell+1}}{\boldsymbol{r}_{\ell}^\top\boldsymbol{r}_{\ell}}$.
  \STATE Update $\boldsymbol{d}_{\ell+1}=\boldsymbol{r}_{\ell+1}+\beta_{\ell}\boldsymbol{d}_{\ell}$.
  \ENDFOR
  \STATE Convert vector $\boldsymbol{x}_{\ell_{\max}}$ into matrix $\boldsymbol{X}$.
% \RETURN Temporal factor matrix $\boldsymbol{X}$.
\end{algorithmic}
}
\end{algorithm}

\subsection{Coefficient Matrix Estimation}

The coefficient matrices $\{\boldsymbol{A}_k\}_{k\in[d]}$ in NoTMF represent temporal correlations of the temporal factors $\{\boldsymbol{x}_t\}_{t\in[T]}$. According to the property of the Kronecker product, one can rewrite the formula such that
\begin{equation}
\begin{aligned}
\sum_{k\in[d]}\boldsymbol{A}_k\boldsymbol{X}\boldsymbol{\Psi}_k^\top=&\boldsymbol{A}_{1}\boldsymbol{X}\boldsymbol{\Psi}_1^\top+\boldsymbol{A}_{2}\boldsymbol{X}\boldsymbol{\Psi}_2^\top+\cdots +\boldsymbol{A}_{d}\boldsymbol{X}\boldsymbol{\Psi}_d^\top \\
=&\begin{bmatrix}
\boldsymbol{A}_1 & \boldsymbol{A}_2 & \cdots & \boldsymbol{A}_d
\end{bmatrix}\underbrace{\begin{bmatrix}
\boldsymbol{X} & \boldsymbol{0} & \cdots & \boldsymbol{0} \\
\boldsymbol{0} & \boldsymbol{X} & \cdots & \boldsymbol{0} \\
\vdots & \vdots & \ddots & \vdots \\
\boldsymbol{0} & \boldsymbol{0} & \cdots & \boldsymbol{X} \\
\end{bmatrix}}_{=\boldsymbol{I}_d\otimes\boldsymbol{X}}\begin{bmatrix}
\boldsymbol{\Psi}_1^\top \\
\boldsymbol{\Psi}_2^\top \\
\vdots \\
\boldsymbol{\Psi}_d^\top \\
\end{bmatrix},
\end{aligned}
\end{equation}
and as a consequence, the subproblem with respect to the coefficient matrices can be formulated as
\begin{equation}
\min_{\boldsymbol{A}}~\frac{\gamma}{2}\bigl\|\boldsymbol{X}\boldsymbol{\Psi}_0^\top-\boldsymbol{A}(\boldsymbol{I}_d\otimes\boldsymbol{X})\boldsymbol{\Psi}^\top\bigr\|_F^2,
\end{equation}
where
\begin{equation*}
\begin{aligned}
\boldsymbol{A}=&\begin{bmatrix}
\boldsymbol{A}_1 & \boldsymbol{A}_2 & \cdots & \boldsymbol{A}_d
\end{bmatrix}\in\mathbb{R}^{R\times (dR)}, \\
\boldsymbol{\Psi}=&\begin{bmatrix}
\boldsymbol{\Psi}_1 & \boldsymbol{\Psi}_2 & \cdots & \boldsymbol{\Psi}_d
\end{bmatrix}\in\mathbb{R}^{(T-d-m)\times (dT)},
\end{aligned}
\end{equation*}
are augmented matrices. As each matrix has $d$ sub-matrices, $\boldsymbol{A}$ and $\boldsymbol{\Psi}$ show $dR$ and $dT$ columns, respectively. Since we build correlations by VAR on the temporal factors, the subproblem would only demonstrate the VAR component. The decision variable of this subproblem is the augmented coefficient matrix $\boldsymbol{A}$. The partial derivative of $f$ with respect to the matrix $\boldsymbol{A}$ is given by
\begin{equation}
\frac{\partial f}{\partial\boldsymbol{A}}=-\gamma\boldsymbol{X}\boldsymbol{\Psi}_0^\top\boldsymbol{\Psi}(\boldsymbol{I}_d\otimes\boldsymbol{X})^\top+\gamma\boldsymbol{A}(\boldsymbol{I}_d\otimes\boldsymbol{X})\boldsymbol{\Psi}^\top\boldsymbol{\Psi}(\boldsymbol{I}_d\otimes\boldsymbol{X})^\top.
\end{equation}

Let $\frac{\partial f}{\partial\boldsymbol{A}}=\boldsymbol{0}$, then we have a least squares solution as
\begin{equation}
\boldsymbol{A}=\boldsymbol{X}\boldsymbol{\Psi}_0^\top\bigl((\boldsymbol{I}_d\otimes\boldsymbol{X})\boldsymbol{\Psi}^\top\bigr)^{\dagger},
\end{equation}
where $\cdot^\dagger$ denotes the Moore-Penrose pseudo-inverse.

\subsection{Solution Algorithm of NoTMF}

As mentioned above, the alternating minimization scheme of NoTMF possesses the closed-form solutions to all three variables $\{\boldsymbol{W},\boldsymbol{X},\boldsymbol{A}\}$. For the spatial factor matrix $\boldsymbol{W}$, it is not difficult to write down the least squares solutions to the column vectors $\{\boldsymbol{w}_n\}_{n\in[N]}$ (see \eqref{solution_to_w}), but one can also compute the approximated solution to the matrix $\boldsymbol{W}$ from \eqref{matrix_equation_w} with conjugate gradient. For the temporal factor matrix $\boldsymbol{X}$, computing from \eqref{matrix_equation_x_vector_form} with least squares is not as efficient as computing from \eqref{matrix_equation_x} with conjugate gradient (see Algorithm~\ref{conj_grad_to_X_algo}). Algorithm~\ref{notmf_algo} summarizes the implementation of NoTMF. In fact, the alternating minimization of NoMTF is the block coordinate minimization with three blocks/variables $\{\boldsymbol{W},\boldsymbol{X},\boldsymbol{A}\}$. One can justify our algorithm decreases the objective function monotonically in the iterative process and the NoTMF algorithm can be converged.

\begin{algorithm}
\caption{NoTMF$(\boldsymbol{Y},\Omega,d,R,\gamma,\rho,m)$}
\label{notmf_algo}
\begin{algorithmic}[1]
 \renewcommand{\algorithmicrequire}{\textbf{Input:}}
 \renewcommand{\algorithmicensure}{\textbf{Output:}}
 \REQUIRE Data matrix $\boldsymbol{Y}$ with the observed index set $\Omega$, order $d\in\mathbb{Z}^{+}$ and season $m\in\mathbb{Z}^{+}$ of the VAR, rank $R\in\mathbb{Z}^+$, and hyperparameters $\{\gamma,\rho\}$.
 \ENSURE Spatial factor matrix $\boldsymbol{W}$, temporal factor matrix $\boldsymbol{X}$, and coefficient matrix $\boldsymbol{A}$.
 \STATE Initialize the variables $\{\boldsymbol{W},\boldsymbol{X},\boldsymbol{A}\}$.
 \STATE Generate temporal operator matrices $\{\boldsymbol{\Psi}_k\}_{t\in[0,d]}$.
 \REPEAT
  \STATE Compute $\boldsymbol{W}$ as the least squares solution (or conjugate gradient).
  \STATE Compute $\boldsymbol{X}$ from \eqref{matrix_equation_x} with conjugate gradient.
  \STATE Compute $\boldsymbol{A}$ by the least squares solution.
\UNTIL{convergence}
% \RETURN $\boldsymbol{W},\boldsymbol{X},\boldsymbol{A}$.
\end{algorithmic}
\end{algorithm}

\subsection{Rolling Forecasting with Dictionary Learning}

\begin{figure}[htbp]
    \centering
    \begin{subfigure}{.78\linewidth}
    \centering
    \includegraphics[width=\textwidth]{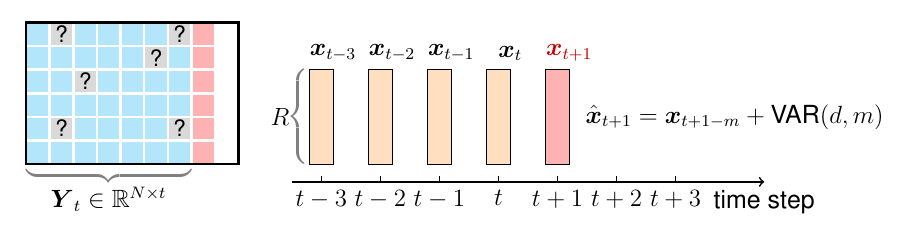}
    \caption{At the $t$-th time step}
    \end{subfigure}
    \begin{subfigure}{.8\linewidth}
    \centering
    \includegraphics[width=\textwidth]{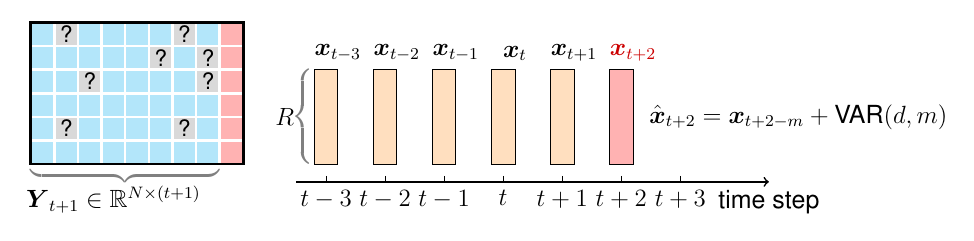}
    \caption{At the $t+1$-th time step}
    \end{subfigure}
    \vspace{0.2cm}
    \caption{Sparse movement speed forecasting with NoTMF in which $\operatorname{VAR}(d,m)$ is the VAR with the order $d$ and the season $m$.}
    \label{notmf_forecast_xt}
\end{figure}

In the following, we introduce the forecasting scheme for NoTMF. For example, Figure~\ref{notmf_forecast_xt} shows how NoTMF works on the single-step movement speed forecasting in the latent spaces. At the $t$-th time step, as shown in Figure~\ref{notmf_forecast_xt}(a), we can first estimate the spatial factor matrix, the temporal factor matrix, and the coefficient matrix on the sparse input data $\boldsymbol{Y}_t\in\mathbb{R}^{N\times t}$ (see Algorithm~\ref{notmf_algo}), then we can forecast the next-step temporal factors as $\hat{\boldsymbol{x}}_{t+1}$ by using seasonal differenced VAR. As we have both spatial factor matrix $\boldsymbol{W}$ and temporal factors $\hat{\boldsymbol{x}}_{t+1}$, it is not hard to generate the future movement speed values as $\hat{\boldsymbol{y}}_{t+1}=\boldsymbol{W}^\top\hat{\boldsymbol{x}}_{t+1}$. Since we already have the spatial factor matrix $\boldsymbol{W}$, we can fix this variable as a dictionary matrix and instantly compute the temporal factor matrix and the coefficient matrix at the $t+1$-th time step, referring to the online dictionary learning as mentioned by \cite{gultekin2018online}. In Figure~\ref{notmf_forecast_xt}(b), we can first use input data $\boldsymbol{Y}_{t+1}\in\mathbb{R}^{N\times (t+1)}$ to update the temporal factor matrix and the coefficient matrix. Then, the seasonal differenced VAR allows one to forecast the next step temporal factors as $\hat{\boldsymbol{x}}_{t+2}$ and generate the future movement speed values as $\hat{\boldsymbol{y}}_{t+2}=\boldsymbol{W}^\top\hat{\boldsymbol{x}}_{t+2}$.

In this work, we consider $\delta$-step forecasting on the sparse movement speed data. Algorithm~\ref{notmf_algo} provides an efficient routine for estimating $\{\boldsymbol{W},\boldsymbol{X},\boldsymbol{A}\}$. We first perform seasonal differencing on temporal factors $\{\boldsymbol{x}_t\}_{t\in[T]}$ as $\dot{\boldsymbol{x}}_{t}^{m}=\boldsymbol{x}_{t}-\boldsymbol{x}_{t-m}$ for all $t\in[m+1, T]$, and then forecast $\dot{\boldsymbol{x}}_{T+1}^{m},\ldots,\dot{\boldsymbol{x}}_{T+\delta}^{m}$ through the seasonal differenced VAR with the time horizon $\delta\in\mathbb{Z}^+$. Finally, we generate the forecasts of movement speed values $\hat{\boldsymbol{y}}_{T+1},\ldots,\hat{\boldsymbol{y}}_{T+\delta}$ sequentially by
\begin{equation}\label{data_point_forecasts}
\hat{\boldsymbol{y}}_{T+1}=\boldsymbol{W}^\top\hat{\boldsymbol{x}}_{T+1},\quad\hat{\boldsymbol{y}}_{T+2}=\boldsymbol{W}^\top\hat{\boldsymbol{x}}_{T+2},\quad\ldots,\quad\hat{\boldsymbol{y}}_{T+\delta}=\boldsymbol{W}^\top\hat{\boldsymbol{x}}_{T+\delta}.
\end{equation}

When performing rolling forecasting on sparse movement speed data, we denote the incremental data at the $\kappa$-th rolling forecasting with the time horizon $\delta$ by $\boldsymbol{Y}_{\kappa}\in\mathbb{R}^{N\times (T+(\kappa-1)\cdot\delta)}$. NoTMF can process the entire data at once, but it is impractical to accommodate data arriving sequentially.  Thus, we take into account the dictionary learning with a fixed spatial factor matrix (also examined by \cite{gultekin2018online}). Such a collection of spatial factors enables the system to transform the incremental movement data to lower-dimensional temporal factors. At each rolling time, we first learn the temporal factor matrix through the conjugate gradient method. Then we update the coefficient matrix accordingly to capture time-evolving patterns. Algorithm~\ref{rolling_forecast_algo} shows the implementation of rolling forecasting on the sparse movement speed data.

\begin{algorithm}
\caption{NoTMF Rolling Forecasting on Sparse Movement Speed Data}
\label{rolling_forecast_algo}
\begin{algorithmic}[1]
 \renewcommand{\algorithmicrequire}{\textbf{Input:}}
 \renewcommand{\algorithmicensure}{\textbf{Output:}}
 \REQUIRE Data matrix $\boldsymbol{Y}\in\mathbb{R}^{N\times T}$ with the observed index set $\Omega$, order $d\in\mathbb{Z}^{+}$ and season $m\in\mathbb{Z}^{+}$ of VAR, rank $R\in\mathbb{Z}^{+}$, $K$ rolling times, and forecasting time horizon $\delta\in\mathbb{Z}^{+}$.
 \ENSURE Movement speed forecasts $\hat{\boldsymbol{Y}}\in\mathbb{R}^{N\times \delta\cdot K}$.
 \STATE Train NoTMF on data matrix $\boldsymbol{Y}$ as shown in Algorithm~\ref{notmf_algo} and return the result:\\ $\boldsymbol{W},\boldsymbol{X},\boldsymbol{A}=\text{NoTMF}(\boldsymbol{Y},\Omega,d,R,\gamma,\rho,m)$.
 \STATE Generate the forecasts $\hat{\boldsymbol{y}}_{T+1},\ldots,\hat{\boldsymbol{y}}_{T+\delta}$ by \eqref{data_point_forecasts} and stack them as the first $\delta$ column vectors of $\hat{\boldsymbol{Y}}$.
  \FOR {$\kappa = 2$ to $K$}
  \STATE Input the incremental movement speed data $\boldsymbol{Y}_{\kappa}\in\mathbb{R}^{N\times (T+(\kappa-1)\cdot\delta)}$ in which the last $\delta$ column vectors are the newly arriving data.
  \STATE Generate temporal operator matrices $\{\boldsymbol{\Psi}_k\}_{k\in[0,d]}$.
  \STATE Compute $\boldsymbol{X}_{\kappa}$ from data $\boldsymbol{Y}_\kappa$ with conjugate gradient.
  \STATE Update the coefficient matrix $\boldsymbol{A}$.
  \STATE Generate the forecasts $\hat{\boldsymbol{y}}_{T+(\kappa-1)\cdot\delta+1},\ldots,\hat{\boldsymbol{y}}_{T+(\kappa-1)\cdot\delta+\delta}$ and stack them to $\hat{\boldsymbol{Y}}$.
  \ENDFOR
% \RETURN $\hat{\boldsymbol{Y}}$.
\end{algorithmic}
\end{algorithm}

\section{Experiments on Uber Movement Data}

Within the context of this study, we conduct an empirical investigation for the NoTMF model through forecasting sparse movement speeds in urban road networks. Specifically, we evaluate NoTMF based on two large-scale movement datasets: the NYC Uber movement speed dataset and the Seattle Uber movement speed dataset. Both datasets are characterized by hourly movement speed values, derived from the trajectories of ridesharing vehicles across various road segments. But if one road segment at a certain hour does not cover at least five unique trips/traces, then the corresponding road-level traffic speed would be ignored. In light of this, both datasets involve a large amount of missing/unobserved values, thereby accentuating the datasets' inherent sparsity.

\subsection{Data Analysis and Experiment Setup}

\begin{figure}[htbp]
    \centering
    \begin{subfigure}{.65\linewidth}
    \centering
    \includegraphics[width=\textwidth]{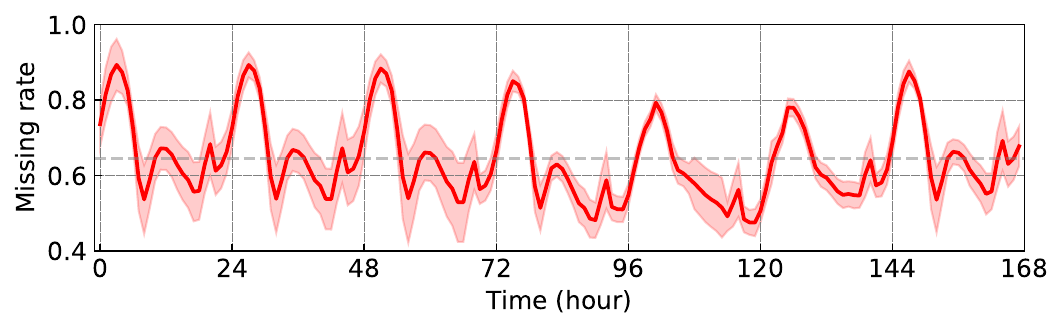}
    \caption{NYC Uber movement speed data.}
    \end{subfigure}
    \begin{subfigure}{.65\linewidth}
    \centering
    \includegraphics[width=\textwidth]{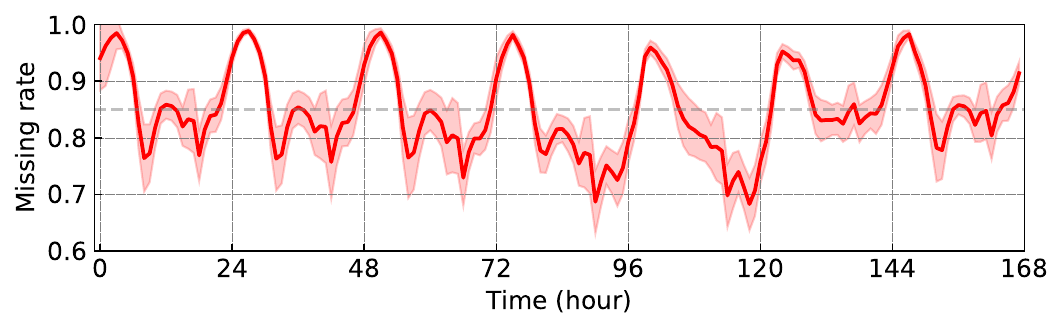}
    \caption{Seattle Uber movement speed data.}
    \end{subfigure}
\caption{The missing rates of Uber movement speed data aggregated per week over the whole year of 2019. The red curve shows the average missing rates over all 52 weeks. The red area shows the standard deviation of missing rates in each hour over 52 weeks. The 168 time steps refer to 168 hours of Tuesday, Wednesday, Thursday, Friday, Saturday, Sunday, and Monday. (a) The dataset has 98,210 road segments, and the overall missing rate is 64.43\%. (b) The dataset has 63,490 road segments, and the overall missing rate is 84.95\%.}
\label{missing_rate_stat}
\end{figure}

The datasets originate from the Uber Movement project, which aims to equip cities with data and analytical tools to navigate and address urban transportation challenges. 
% The project provides data and tools for cities to understand and address urban transportation problems and challenges in a data-driven fashion.
The project's movement speed data, indicative of average speeds across specified urban road segments, represent the high-dimensional and sparse nature of traffic data encountered in real-world settings.
% Uber movement speed data measure the average speed on a given road segment in urban areas for each hour of each day over the study period, and they are representative examples for demonstrating both high-dimensionality and sparsity issues in real-world traffic data.
Figure~\ref{missing_rate_stat} shows two cases of Uber movement speed data in NYC and Seattle, USA, respectively. The high-dimensional and sparse data inevitably result in difficulties in analyzing traffic states or supporting data-driven city planning and decision-making. 

\begin{figure}[h!]
\centering
\includegraphics[width = 0.6\textwidth]{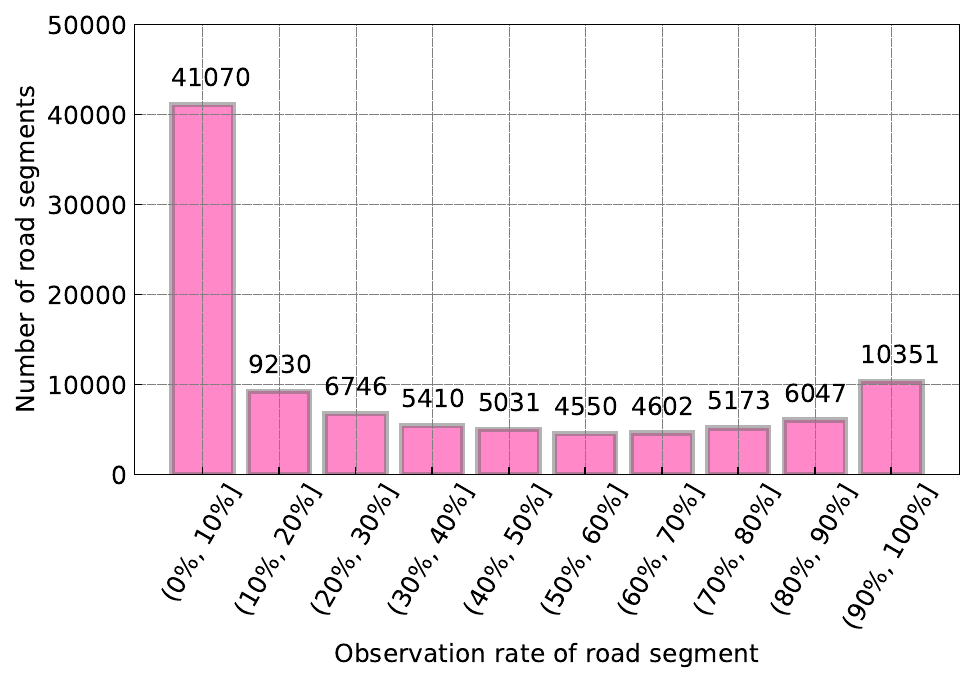}
\caption{Histogram of observation rate of road segment in the NYC Uber movement speed dataset. Only a small fraction of road segments have an observation rate greater than 50\%, i.e., $30723/98210\approx31\%$. For the observation rates greater than 20\% and 80\%, there are about 49\% and 17\% of road segments, respectively.}
\label{road_segment_obs_ratio}
\end{figure}

\begin{figure}[h!]
    \centering
    \includegraphics[width = 0.75\textwidth]{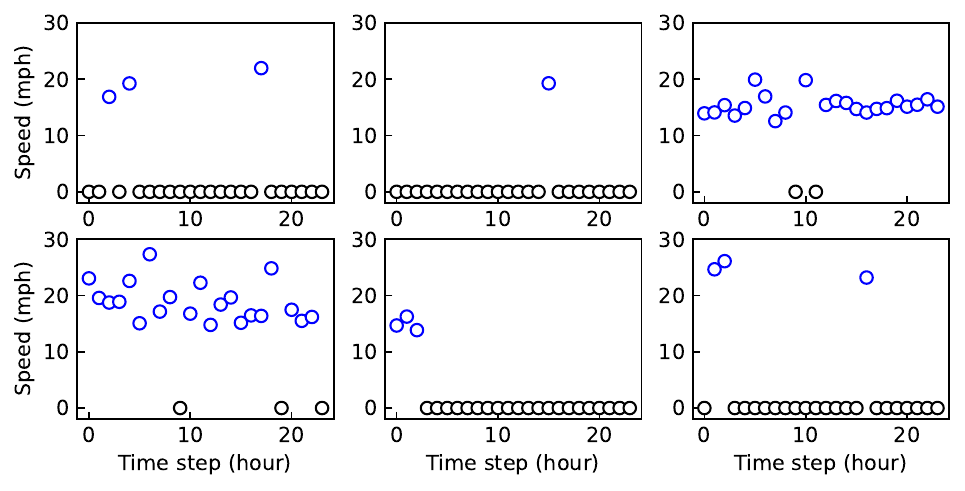}
    \caption{Movement speed of 6 road segments of January 1, 2019 (24 hours) in NYC. Blue points indicate the observed speed from the movement dataset, while black points indicate missing values (set to 0).}
    \label{NYC_missing_data_scatter}
\end{figure}

For our experiments, we choose the data collected over the first ten weeks of 2019 from 98,210 road segments in NYC, comprising a dataset matrix size of $98210\times 1680$. Due to the insufficient sampling of ridesharing vehicles in urban road networks, the dataset contains 66.56\% missing values. 
It can be seen that the time-evolving missing rates have periodicity patterns. At midnight, the missing rate can even reach $\sim$90\%. 
This is due to limited ridesharing vehicles on the road network during nights. At some specific hours, only a small fraction of road segments have speed observations. Similarly, the Seattle Uber movement speed dataset covers the hourly movement speed data from 63,490 road segments during the first ten weeks of 2019. The data matrix is of size $63490\times 1680$ and contains 87.35\% missing values, showing that the sparsity issue in this dataset is more challenging to handle than the NYC Uber movement speed dataset. Such sparsity underscores the critical need for sophisticated forecasting models capable of navigating these data gaps effectively.

% \begin{remark}
% It is important to note that traditional matrix factorization techniques struggle with column-wise missing data, which is a typical issue pronounced in traffic data, especially during midnight hours when sparsity peaks, as shown in Figure~\ref{missing_rate_stat}. The TMF framework, therefore, represents a significant advancement, enabling more accurate estimation by correlating data columns along the temporal dimension and addressing the challenges posed by data sparsity.
% \end{remark}

Figure~\ref{road_segment_obs_ratio} shows the complementary histogram of the percentage of observed values in the NYC dataset. As can be seen, a large fraction of road segments only have very limited observations. This is especially interesting, as it shows that half of the road segments have less than 20\% movement speed observations. Only 17\% road segments have more than 80\% movement speed observations. We can also see examples of incomplete movement speed observations from Figure~\ref{NYC_missing_data_scatter}.

In our experiment, we evaluate the model with 8-week data (from January $1$st to February $25$th) as the training set, one-week data (February $26$th to March $4$th) for validation, and one-week data (March $5$th to March $11$th) as the test set. We use rolling forecasting with time horizons $\delta=1,2,3,6$, corresponding to $\delta$-hour-ahead forecasting. 
To confirm the importance of seasonal differencing, we evaluate the proposed NoTMF model and baseline models in the TMF framework. We consider the following TMF models in the literature:

\begin{itemize}
\item \textbf{TRMF} \citep{yu2016temporal}: TRMF achieves temporal modeling on latent temporal factors by applying a univariate autoregressive process.
\item \textbf{BTMF} \citep{chen2022bayesian}: BTMF is a fully Bayesian TMF model with Gaussian assumption, which is solved by using the Markov chain Monte Carlo (MCMC) algorithm. This model has been empirically demonstrated to be state-of-the-art against some baseline models (e.g., matrix/tensor factorization) for both time series imputation and forecasting on sparse data.
\item \textbf{BTRMF} \citep{chen2022bayesian}: Bayesian TRMF is a fully Bayesian treatment for the TRMF model with Gaussian assumption.
\item \textbf{CTNNM} \citep{liu2022recovery, liu2022time}: Circulant tensor nuclear norm minimization that is implemented by the fast Fourier transform.
\end{itemize}

\subsection{Forecasting Performance}

\begin{figure}
    \centering
    \includegraphics[width = .6\linewidth]{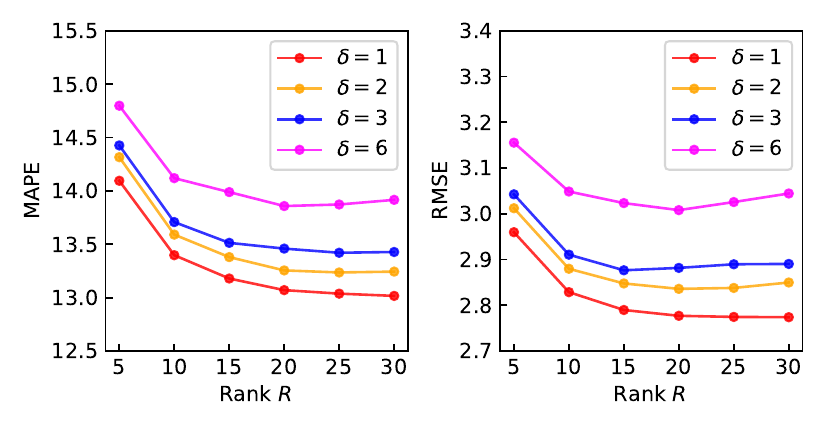}
    \caption{Performance of NoTMF with season-168 differencing (i.e., $m=168$) and order $d=6$ on the NYC dataset. The validated parameters are $\gamma=1$ and $\rho=5$.}
    \label{notmf_forecast_performance}
\end{figure}

We evaluate the task of traffic time series forecasting in the presence of missing data on the NYC and Seattle Uber movement speed datasets. The preliminary experiment evaluates the NoTMF model with different ranks $R=5,10,15,20,25,30$. To set the hyperparameters $\{\gamma,\rho\}$, we first prescribe a collection of hyperparameters $\gamma\in\{10^2,10^1,10^0,10^{-1},10^{-2}\}$ and $\rho\in\{10\gamma,5\gamma,\gamma,5\times10^{-1}\gamma,10^{-1}\gamma\}$ and evaluate the NoTMF with these settings on the training and validation sets. Then, we find the best hyperparameter pair for testing the model. In this case, the hyperparameters of NoTMF are properly given by $\gamma=1$ and $\rho=5$. As shown in Figure~\ref{notmf_forecast_performance}, NoTMF with a larger rank essentially provides higher accuracy but such improvement becomes marginal when $R>15$. In the following experiments, we choose $R=10$ for a good balance between performance and computational cost. 

\begin{table*}
\centering
\caption{Forecasting performance (in MAPE/RMSE) on the test set of the NYC movement speed dataset. The rolling forecasting tasks include different time horizons, i.e., $\delta=1,2,3,6$. We consider the rank as $R=10$. The best $(\gamma,\rho)$ is found on the validation set to be $(1,5)$ for NoTMF and TRMF. Since CTNNM does not have the order $d$, we only present the forecasting results along the first row at each time horizon. Note that the best test results are highlighted in bold fonts.}
\label{notmf_forecasting_table}
\begin{tabular}{l|c|c|c|c|c|c|c} 
\toprule
$\delta$    & $d$ & \begin{tabular}[c]{@{}c@{}}NoTMF \\($m=24$)\end{tabular} & \begin{tabular}[c]{@{}c@{}}NoTMF\\~($m=168$)\end{tabular} & TRMF & BTMF & BTRMF & CTNNM \\ 
\midrule
% \cmidrule{1-2}\cmidrule(l){3-6}
\multirow{4}{*}{1} & 1   & 13.63/2.88 & \textbf{13.53}/\textbf{2.86} 
% & \textbf{13.45}/\textbf{2.85} %& 13.74/2.90 
& 14.50/3.12  & 14.94/3.13 & 15.93/3.33 & 15.79/3.12 \\
 & 2   & 13.47/\textbf{2.84} & \textbf{13.41}/\textbf{2.84} 
 % & 13.42/\textbf{2.84} %& 13.53/2.85 
 & 14.14/3.05 & 15.70/3.41 & 15.90/3.35 & -/- \\
 & 3   & 13.46/2.84 & \textbf{13.39}/\textbf{2.83} 
 % & 13.43/2.84 %& 13.47/\textbf{2.83} 
 & 13.87/2.96  & 15.80/3.34 & 16.08/3.43 & -/- \\
 & 6   & 13.41/\textbf{2.83} & \textbf{13.39}/\textbf{2.83} 
 % & 13.41/\textbf{2.83} %& 13.40/\textbf{2.83} 
 & 14.00/2.98  & 15.45/3.27 & 16.26/3.48 & -/- \\
\midrule
\multirow{4}{*}{2} & 1   & 13.91/2.96 & \textbf{13.76}/\textbf{2.94} 
% & \textbf{13.70}/\textbf{2.92} %& 14.24/3.00 
& 15.85/3.43 & 15.33/3.21 & 16.85/3.56 & 15.91/3.14 \\
 & 2   & 13.77/2.92 & \textbf{13.63}/\textbf{2.89} 
 % & 13.72/2.92 %& 13.87/2.91 
 & 15.04/3.31 & 15.87/3.32 & 17.27/3.71 & -/- \\
 & 3   & 13.72/2.91 & \textbf{13.61}/\textbf{2.89} 
 % & 13.73/2.92 %& 13.81/\textbf{2.89} 
 & 15.25/3.36 & 15.69/3.33 & 17.24/3.74 & -/- \\
 & 6   & 13.59/\textbf{2.87} & \textbf{13.57}/2.88 
 % & 13.68/2.91 %& 13.63/\textbf{2.86}   
 & 14.92/3.24 & 15.91/3.39 & 18.18/3.97 & -/- \\
\midrule
\multirow{4}{*}{3} & 1   & 14.30/3.05 & \textbf{14.06}/\textbf{3.02} 
% & \textbf{14.02}/\textbf{3.00} %& 14.81/3.12 
& 17.52/3.83 & 15.86/3.32 & 18.61/3.91 & 16.02/3.16 \\
 & 2   & 14.01/2.98 & \textbf{13.84}/\textbf{2.94} 
 % & 13.96/2.98 %& 14.26/2.98 
 & 17.32/4.00 & 16.30/3.40 & 18.90/4.10 & -/- \\
 & 3   & 13.95/2.97 & \textbf{13.79}/\textbf{2.93} 
 % & 13.98/2.98 %& 14.04/2.94 
 & 16.91/3.71 & 16.56/3.49 & 18.68/4.05 & -/- \\
 & 6   & 13.78/\textbf{2.92} & \textbf{13.73}/\textbf{2.92} 
 % & 13.91/2.96 %& 13.94/\textbf{2.92}   
 & 16.72/3.65 & 15.49/3.27 & 20.45/4.66 & -/- \\
\midrule
\multirow{4}{*}{6} & 1   & \textbf{14.61}/\textbf{3.11} & 14.67/3.20  
% & 14.98/3.32 %& 15.41/3.21 
& 21.20/4.70 & 15.99/3.32 & 22.40/4.69 & 15.96/3.15 \\
 & 2   & \textbf{14.30}/\textbf{3.03} & 14.33/3.09     
 % & 14.90/3.28 %& 14.85/3.07 
 & 20.87/5.01 & 16.04/3.33 & 23.56/5.63 & -/- \\
 & 3   & \textbf{14.26}/\textbf{3.03} & 14.28/3.09  
 % & 14.86/3.26 %& 14.57/\textbf{3.01} 
 & 20.08/4.65 & 15.67/3.28 & 24.27/5.72 & -/- \\
 & 6   & \textbf{14.06}/\textbf{2.97} & 14.16/3.06 
 % & 14.80/3.23 %& 14.47/3.00   
 & 20.40/4.35 & 16.38/3.50 & 26.34/6.60 & -/- \\
\bottomrule
\end{tabular}
\end{table*}

Table~\ref{notmf_forecasting_table} shows the forecasting performance of NoTMF and baseline models on the test set from the NYC dataset. %The forecasting approaches include NoTMF and the baseline models.  
We summarize the following findings from the results:
\begin{itemize}
\item With the increase in forecasting time horizons, the forecasting errors of all models increase. For each time horizon, as the order increases, the forecasting performance of NoTMF and TRMF is improved. % Observing these models with different ranks, the forecasting approaches with rank 10 outperform those with rank 5.
\item CTNNM, by its definition, is capable of global time series trend modeling, therefore the forecasting performance with the increase of time horizons does not have a significant accuracy reduction. The overall performance of CTNNM is as competitive as BTMF.
\item The NoTMF models with different differencing operations demonstrate a significant improvement over TRMF in terms of forecasting accuracy. In contrast to the univariate autoregressive process, there is a clear benefit from temporal modeling with a multivariate VAR process.
\item On this dataset, we have two choices for setting the season. One is $m=24$, corresponding to the daily differencing, while another is $m=168$, corresponding to the weekly differencing. For both settings, NoTMF can achieve competitively accurate forecasts.
\end{itemize}

\begin{figure}[!t]
    \centering
    \includegraphics[width = 0.5\textwidth]{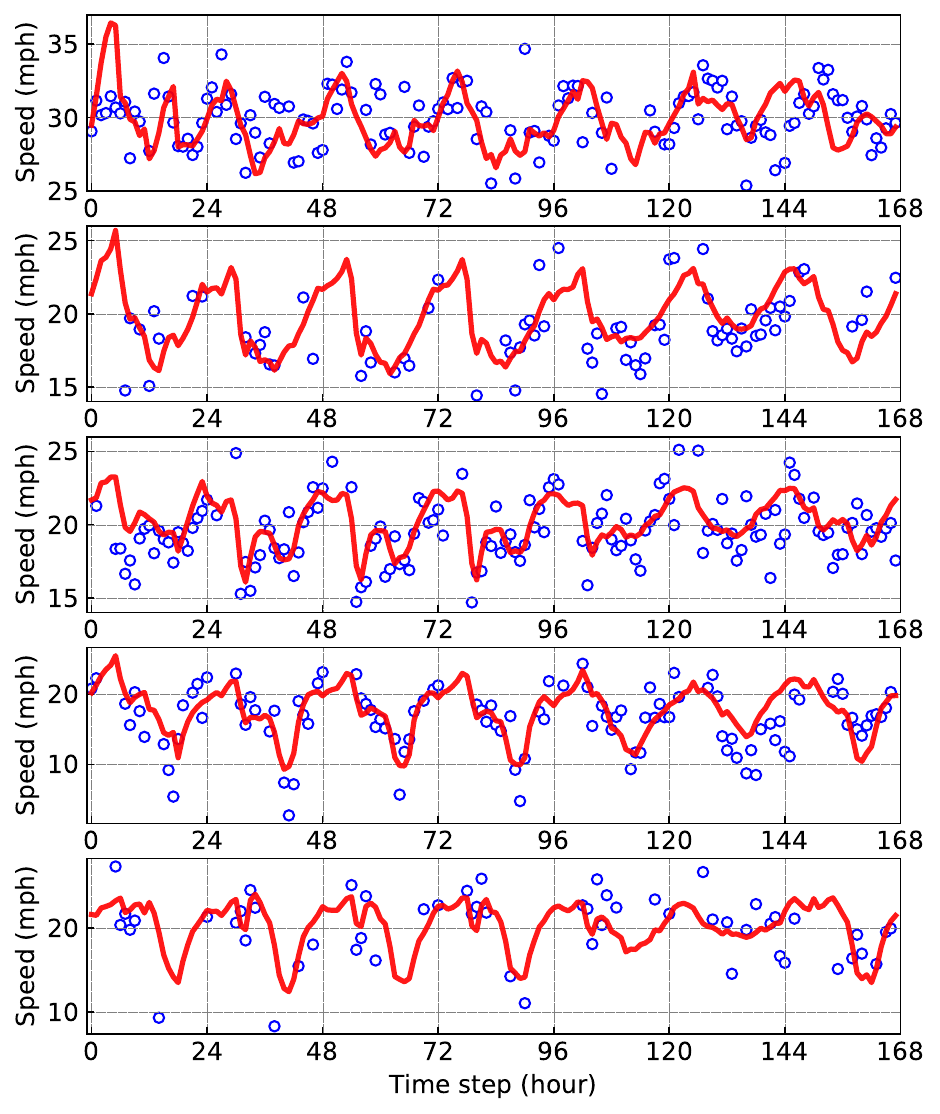}
    \caption{Five examples from the test set (corresponding to five road segments) for showing the forecasting results of the NoTMF model with time horizon $\delta=6$. The red curves indicate the forecasts in the testing week, while the blue scatters indicate the ground truth speed data.}
    \label{notmf_forecasts_delta_6}
\end{figure}

Figure~\ref{notmf_forecasts_delta_6} shows the predicted time series and the partially observed speed values of some road segments. It demonstrates that the forecasts produced by NoTMF are consistent with the temporal patterns underlying partially observed time series. Thus, the NoTMF model can extract implicit temporal patterns (see Figure~\ref{temporal_factor} for instance) from partially observed data and in the meanwhile perform forecasting.

On the Seattle dataset, we perform forecasting and compare the proposed NoTMF with some baseline models as shown in Table~\ref{notmf_forecasting_table_seattle}. Of these results, it seems that VAR in the NoTMF and BTMF models helps produce reliable forecasting performance with the increase of forecasting time horizons, while the univariate autoregression in the TRMF and BTRMF models fails to maintain the performance of the same level as the increase of forecasting time horizons. Notably, NoTMF consistently outperforms other models as shown in Table~\ref{notmf_forecasting_table_seattle}. We set NoTMF with both daily differencing (i.e., $m=24$) and weekly differencing (i.e., $m=168$), and it shows that the weekly differencing is superior to the daily differencing. Therefore, proper seasonal differencing in NoTMF is important for improving the forecasting performance.

\begin{table*}
\centering
\caption{Forecasting performance (MAPE/RMSE) on the test set of the Seattle movement speed dataset. The rolling forecasting tasks include different time horizons, i.e., $\delta=1,2,3,6$. We consider the rank as $R=10$. The best $(\gamma,\rho)$ is found on the validation set to be $(1,5)$ for NoTMF and TRMF. Since CTNNM does not have the order $d$, we only present the forecasting results along the first row at each time horizon. Note that the best test results are highlighted in bold fonts.}
\label{notmf_forecasting_table_seattle}
\begin{tabular}{l|l|c|c|c|c|c|c} 
\toprule
$\delta$    & $d$ & \begin{tabular}[c]{@{}c@{}}NoTMF \\($m=24$)\end{tabular} & \begin{tabular}[c]{@{}c@{}}NoTMF\\~($m=168$)\end{tabular} & TRMF & BTMF & BTRMF & CTNNM \\ 
\midrule
% \cmidrule{1-2}\cmidrule(l){3-6}
\multirow{4}{*}{1}
& 1 & 10.45/3.32 & \textbf{10.26}/\textbf{3.22} 
% & \textbf{10.26}/\textbf{3.21} %& 10.70/3.37 
& 11.58/3.79 & 12.23/3.89 & 12.52/4.01 & 12.74/3.83 \\
& 2 & 10.53/3.34 & \textbf{10.29}/\textbf{3.23} 
% & \textbf{10.23}/\textbf{3.21} %& 10.47/3.28 
& 10.92/3.51 & 12.95/4.18 & 13.16/4.31 & -/- \\
& 3 & 10.42/3.30 & \textbf{10.30}/\textbf{3.22} 
% & \textbf{10.25}/\textbf{3.21} %& 10.40/3.27 
& 10.86/3.47 & 12.96/4.22 & 13.89/4.64 & -/- \\
& 6 & 10.50/3.32 & \textbf{10.21}/\textbf{3.21} 
% & 10.27/3.22 %& 10.32/3.24 
& 10.99/3.51 & 12.91/4.18 & 13.90/4.67 & -/- \\
\midrule
\multirow{4}{*}{2} 
& 1 & 10.90/3.55 & \textbf{10.32}/\textbf{3.25} 
% & \textbf{10.25}/\textbf{3.23} %& 11.11/3.52 
& 12.07/4.02 & 12.74/4.06 & 13.31/4.32 & 12.97/3.90 \\
& 2 & 10.90/3.52 & \textbf{10.31}/\textbf{3.24} 
% & \textbf{10.25}/\textbf{3.23} %& 10.99/3.48 
& 12.59/4.24 & 13.68/4.45 & 13.44/4.43 & -/- \\
& 3 & 10.81/3.49 & \textbf{10.31}/\textbf{3.24} 
% & \textbf{10.27}/\textbf{3.23} %& 10.75/3.40 
& 12.01/3.96 & 13.55/4.46 & 13.66/4.56 & -/- \\
& 6 & 10.57/3.38 & \textbf{10.25}/\textbf{3.23} 
% & 10.27/\textbf{3.23} %& 10.69/3.38 
& 12.18/3.98 & 13.56/4.42 & 14.67/4.92 & -/- \\
\midrule
\multirow{4}{*}{3} 
& 1 & 11.27/3.71 & \textbf{10.41}/\textbf{3.29} 
% & \textbf{10.41}/\textbf{3.29} %& 11.80/3.77 
& 13.47/4.62 & 13.16/4.15 & 14.01/4.52 & 13.36/3.97 \\
& 2 & 11.26/3.71 & \textbf{10.30}/\textbf{3.27} 
% & 10.34/\textbf{3.27} %& 11.93/3.83 
& 14.48/5.19 & 13.63/4.37 & 14.39/4.76 & -/- \\
& 3 & 11.11/3.62 & \textbf{10.35}/\textbf{3.28} 
% & 10.38/\textbf{3.28} %& 11.62/3.70 
& 14.04/4.83 & 13.76/4.42 & 14.67/4.84 & -/- \\
& 6 & 10.96/3.55 & \textbf{10.30}/\textbf{3.26} 
% & \textbf{10.30}/\textbf{3.26} %& 11.07/3.52 
& 13.32/4.51 & 13.28/4.29 & 15.64/5.31 & -/- \\
\midrule
\multirow{4}{*}{6} 
& 1 & 11.88/3.97 & \textbf{10.63}/\textbf{3.43} 
% & \textbf{10.60}/\textbf{3.42} %& 13.17/4.18 
& 15.59/5.32 & 13.63/4.30 & 16.39/5.28 & 13.82/4.04 \\
& 2 & 11.58/3.83 & \textbf{10.55}/\textbf{3.40} 
% & 10.56/\textbf{3.40} %& 12.61/3.99 
& 18.66/7.20 & 13.27/4.19 & 16.77/5.58 & -/- \\
& 3 & 11.54/3.81 & \textbf{10.57}/\textbf{3.39} 
% & \textbf{10.53}/\textbf{3.38} %& 12.22/3.85 
& 17.94/6.32 & 13.88/4.36 & 17.35/5.70 & -/- \\
& 6 & 11.27/3.70 & \textbf{10.53}/\textbf{3.35} 
% & \textbf{10.50}/\textbf{3.35} %& 11.37/3.59 
& 15.12/5.24 & 13.30/4.24 & 16.63/5.62 & -/- \\
\bottomrule
\end{tabular}
\end{table*}

\begin{figure}[htbp]
    \centering
    \begin{subfigure}{.35\linewidth}
    \centering
    \includegraphics[width=\textwidth]{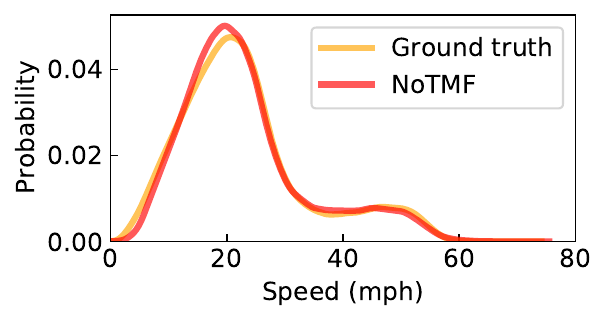}
    \caption{Missing rate between 0 and 10\%; 12,369 road segments.}
    \end{subfigure}
    \begin{subfigure}{.35\linewidth}
    \centering
    \includegraphics[width=\textwidth]{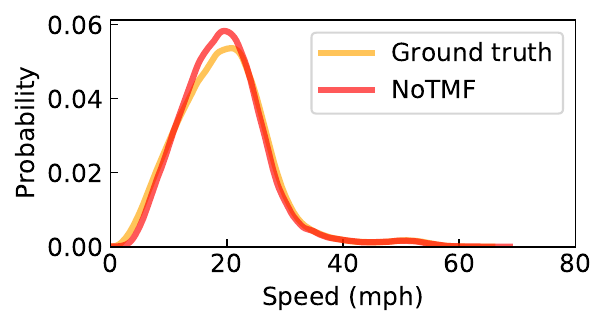}
    \caption{Missing rate between 10\% and 20\%; 7,833 road segments.}
    \end{subfigure}
    \begin{subfigure}{.35\linewidth}
    \centering
    \includegraphics[width=\textwidth]{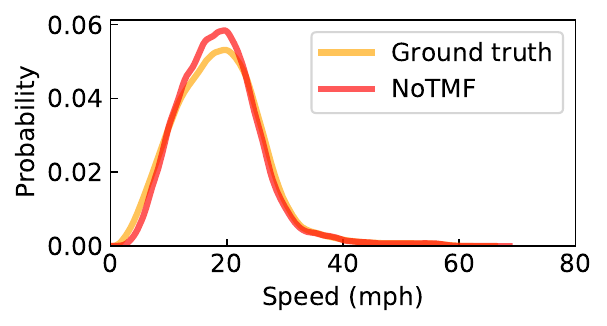}
    \caption{Missing rate between 20\% and 30\%; 6,229 road segments.}
    \end{subfigure}
    \begin{subfigure}{.35\linewidth}
    \centering
    \includegraphics[width=\textwidth]{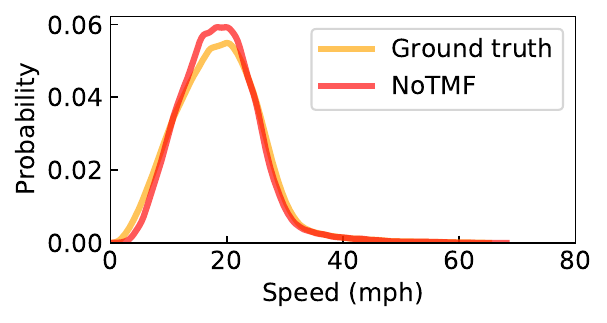}
    \caption{Missing rate between 30\% and 40\%; 5,373 road segments.}
    \end{subfigure}
    \begin{subfigure}{.35\linewidth}
    \centering
    \includegraphics[width=\textwidth]{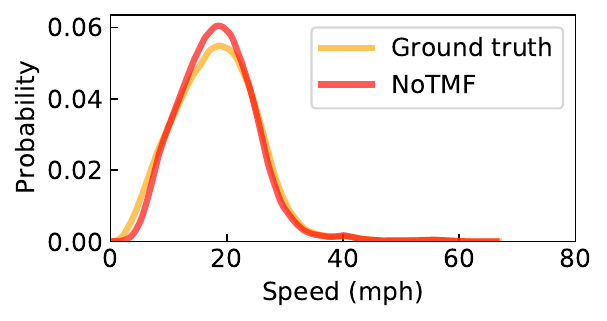}
    \caption{Missing rate between 40\% and 50\%; 5,608 road segments.}
    \end{subfigure}
    \begin{subfigure}{.35\linewidth}
    \centering
    \includegraphics[width=\textwidth]{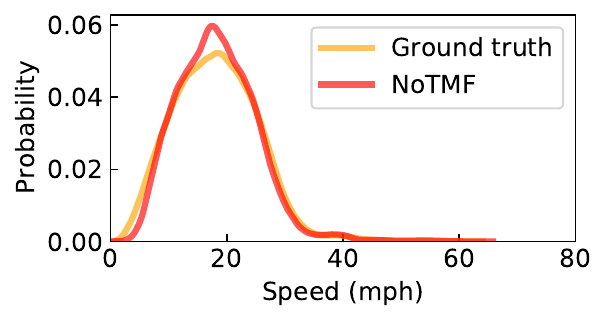}
    \caption{Missing rate between 50\% and 60\%; 4,981 road segments.}
    \end{subfigure}
    \begin{subfigure}{.35\linewidth}
    \centering
    \includegraphics[width=\textwidth]{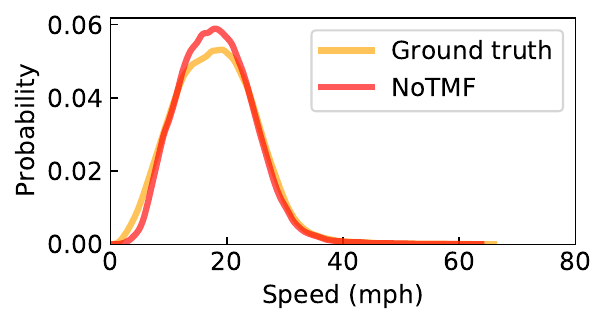}
    \caption{Missing rate between 60\% and 70\%; 5,833 road segments.}
    \end{subfigure}
    \begin{subfigure}{.35\linewidth}
    \centering
    \includegraphics[width=\textwidth]{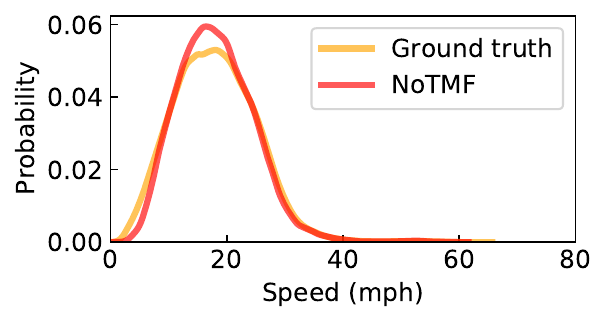}
    \caption{Missing rate between 70\% and 80\%; 6,633 road segments.}
    \end{subfigure}
    \begin{subfigure}{.35\linewidth}
    \centering
    \includegraphics[width=\textwidth]{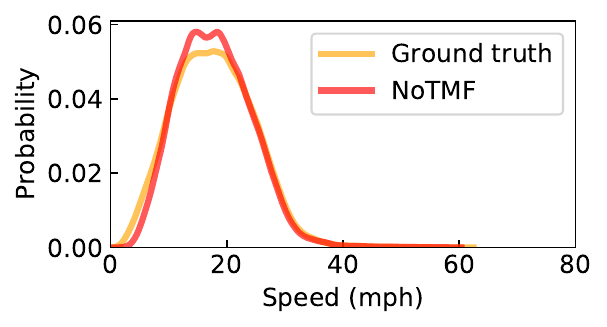}
    \caption{Missing rate between 80\% and 90\%; 9,804 road segments.}
    \end{subfigure}
    \begin{subfigure}{.35\linewidth}
    \centering
    \includegraphics[width=\textwidth]{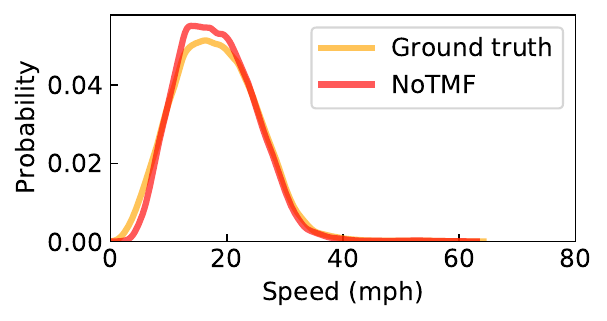}
    \caption{Missing rate between 90\% and 100\%; 33,547 road segments.}
    \end{subfigure}
\caption{The histogram of ground truth data and forecasts achieved by NoTMF with $\delta=6$ and $d=6$ in the test set of the NYC dataset. %The missing rate is the ratio of missing values of road segments in the test set.
}
\label{nyc_forecasts_hist}
\end{figure}

\begin{figure}[htbp]
    \centering
    \begin{subfigure}{.35\linewidth}
    \centering
    \includegraphics[width=\textwidth]{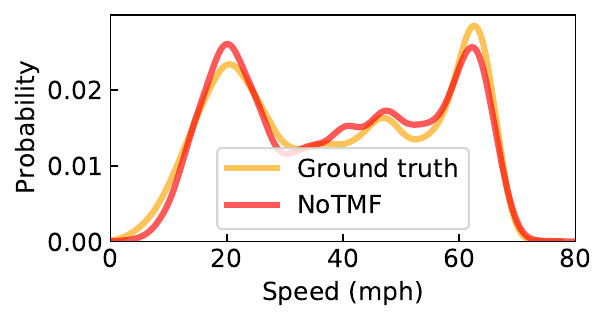}
    \caption{Missing rate between 0 and 10\%; 978 road segments.}
    \end{subfigure}
    \begin{subfigure}{.35\linewidth}
    \centering
    \includegraphics[width=\textwidth]{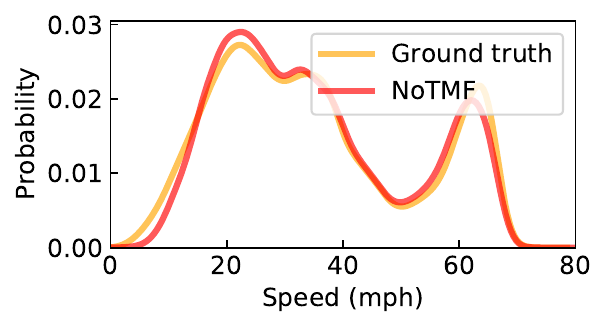}
    \caption{Missing rate between 10\% and 20\%; 1,557 road segments.}
    \end{subfigure}
    \begin{subfigure}{.35\linewidth}
    \centering
    \includegraphics[width=\textwidth]{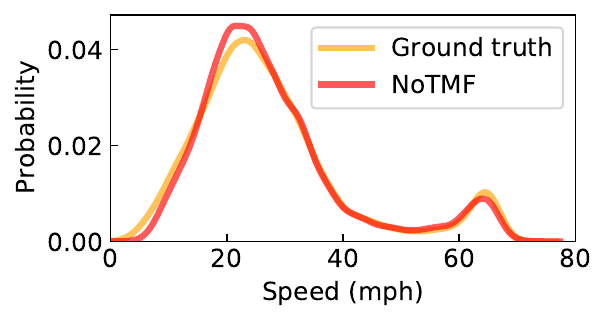}
    \caption{Missing rate between 20\% and 30\%; 1,476 road segments.}
    \end{subfigure}
    \begin{subfigure}{.35\linewidth}
    \centering
    \includegraphics[width=\textwidth]{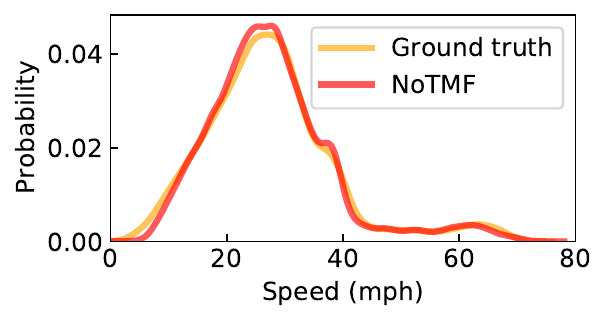}
    \caption{Missing rate between 30\% and 40\%; 1,738 road segments.}
    \end{subfigure}
    \begin{subfigure}{.35\linewidth}
    \centering
    \includegraphics[width=\textwidth]{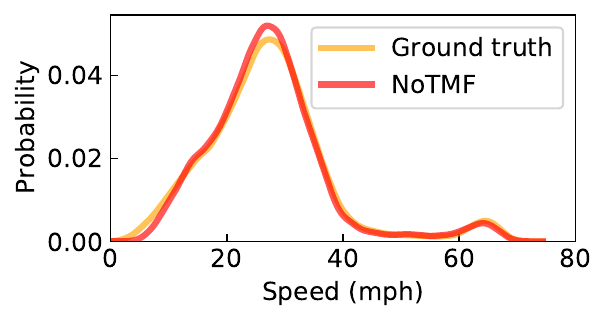}
    \caption{Missing rate between 40\% and 50\%; 1,849 road segments.}
    \end{subfigure}
    \begin{subfigure}{.35\linewidth}
    \centering
    \includegraphics[width=\textwidth]{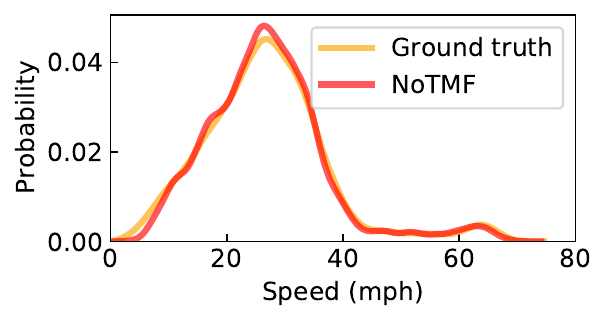}
    \caption{Missing rate between 50\% and 60\%; 1,649 road segments.}
    \end{subfigure}
    \begin{subfigure}{.35\linewidth}
    \centering
    \includegraphics[width=\textwidth]{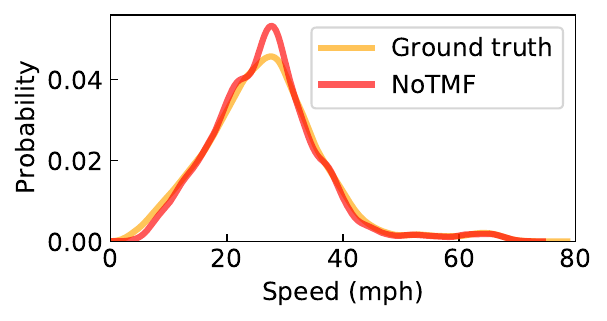}
    \caption{Missing rate between 60\% and 70\%; 2,052 road segments.}
    \end{subfigure}
    \begin{subfigure}{.35\linewidth}
    \centering
    \includegraphics[width=\textwidth]{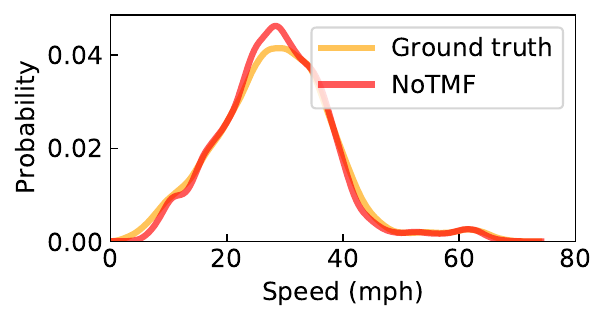}
    \caption{Missing rate between 70\% and 80\%; 3,225 road segments.}
    \end{subfigure}
    \begin{subfigure}{.35\linewidth}
    \centering
    \includegraphics[width=\textwidth]{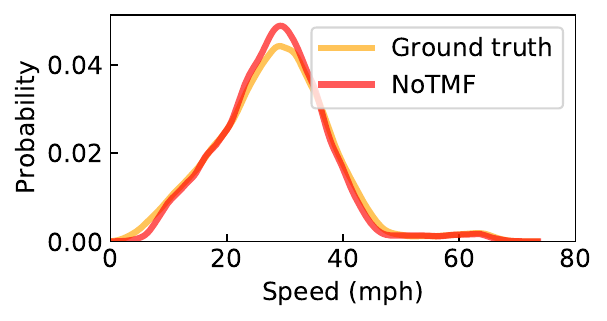}
    \caption{Missing rate between 80\% and 90\%; 6,392 road segments.}
    \end{subfigure}
    \begin{subfigure}{.35\linewidth}
    \centering
    \includegraphics[width=\textwidth]{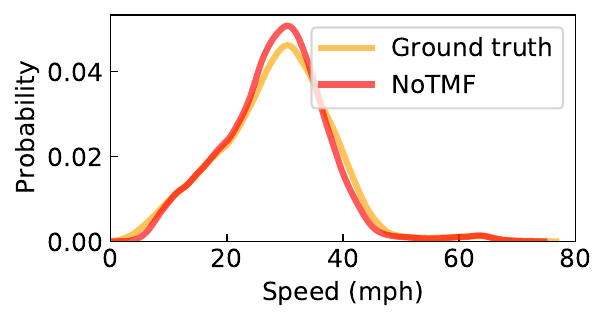}
    \caption{Missing rate between 90\% and 100\%; 42,574 road segments.}
    \end{subfigure}
\caption{The histogram of ground truth data and forecasts achieved by NoTMF with $\delta=6$ and $d=6$ in the test set of the Seattle dataset. %The missing rate is the ratio of missing values of road segments in the test set.
}
\label{seattle_forecasts_hist}
\end{figure}

Figure~\ref{nyc_forecasts_hist} and \ref{seattle_forecasts_hist} show the speed distribution of the ground truth data versus the forecasts achieved by NoTMF on NYC and Seattle datasets, respectively. For each missing rate range, we group the speed observations of the road segments whose missing rate of the test set is in that range. For road segments with relatively lower missing rates, e.g., $(0, 10\%]$, the histograms present two peaks. For example on the Seattle dataset (see Figure~\ref{seattle_forecasts_hist}(a-b)), one peak is around the speed of 20~mph, while another is around the speed of 60~mph. This implies that the road segments with relatively complete speed observations exhibit both lower speeds of congested traffic (possibly during peak hours) and higher speeds of free-flow traffic (possibly during off-peak hours), revealing the bi-modal traffic states. Of these results, NoTMF can accurately forecast both congested and free-flow traffic speeds. %With the increase in missing rates, the phenomena of bi-mode traffic states are weakened. 
As a whole, we can summarize from Figure~\ref{nyc_forecasts_hist} and \ref{seattle_forecasts_hist} that the forecasts produced by NoTMF are accurate when compared to the ground truth data.

\subsection{Nonstationarity Analysis}

\begin{figure}
    \centering
    \includegraphics[width = 0.8\textwidth]{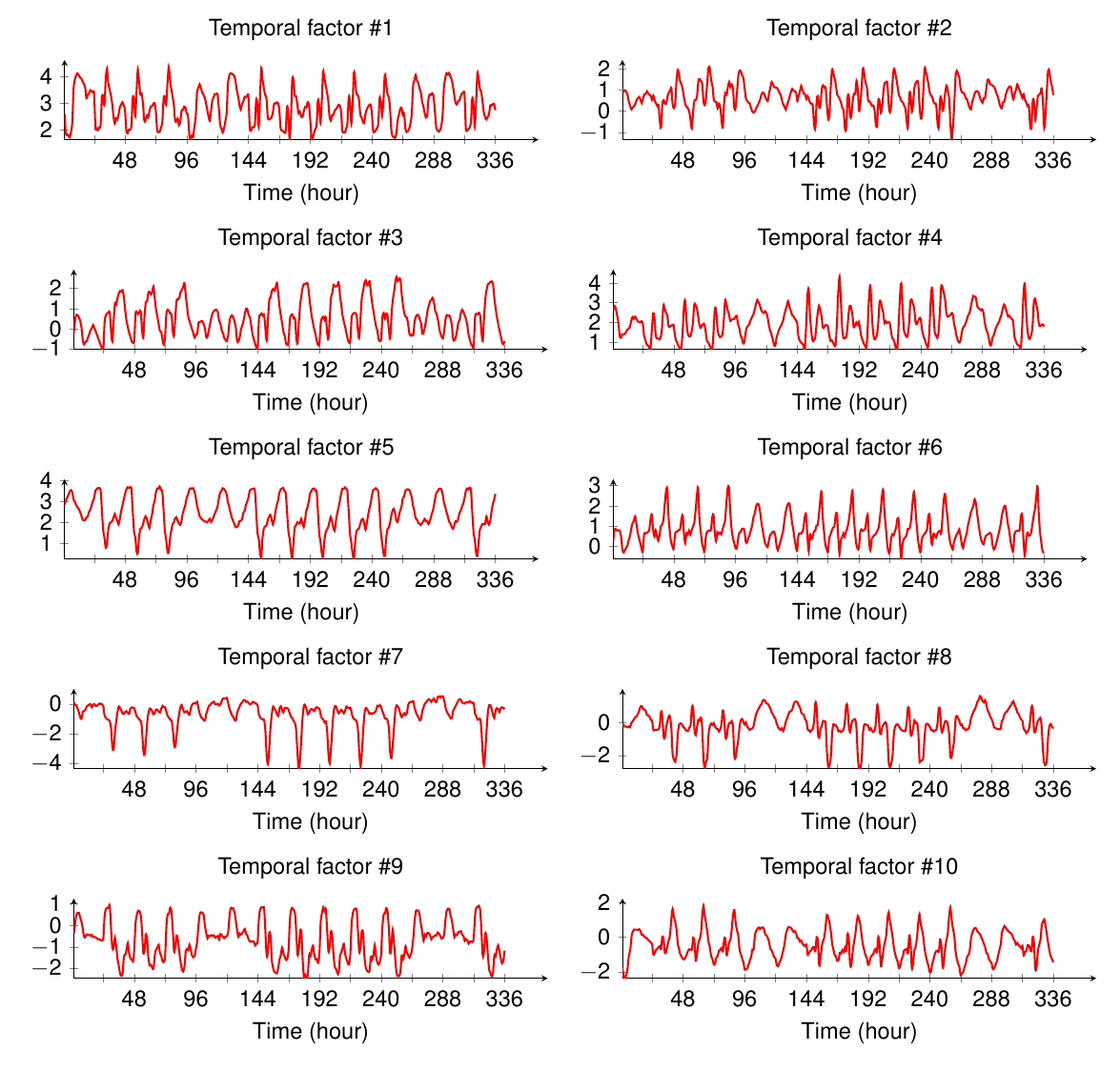}
    \caption{Temporal factors of NoTMF model ($R=10$) on NYC movement speed data. The order and season of NoTMF are set as $d=1$ and $m=24$, respectively. The subfigures only show the temporal factors in the first two weeks.}
    \label{temporal_factor}
\end{figure}

Nonstationarity is an important characteristic in real-world traffic time series data. Figure~\ref{temporal_factor} illustrates some temporal factors of NoTMF with the following setting: $d=1$, $R=10$, and $m=24$. %As can be seen, there are $10$ temporal factors for showing low-dimensional temporal patterns. 
As can be seen, temporal factors \#3, \#5, \#6, \#7, \#8, and \#9 show clear seasonality and periodicity. The long-term season is associated with the week, i.e., $7\times 24=168$ time steps (hours), while the short-term season is associated with the day, i.e., $24$ time steps (hours). For other temporal factors, they also show clear trends with weak seasonality and periodicity. Therefore, making use of seasonal differencing in nonstationary traffic state data can benefit the forecasting performance (as noted in Table~\ref{notmf_forecasting_table} and \ref{notmf_forecasting_table_seattle}). Our results further demonstrate the effectiveness of seasonal differencing in characterizing temporal process in the factor matrix $\boldsymbol{X}$.

\begin{figure}[htbp]
    \centering
    \begin{subfigure}{.7\linewidth}
    \centering
    \includegraphics[width=\textwidth]{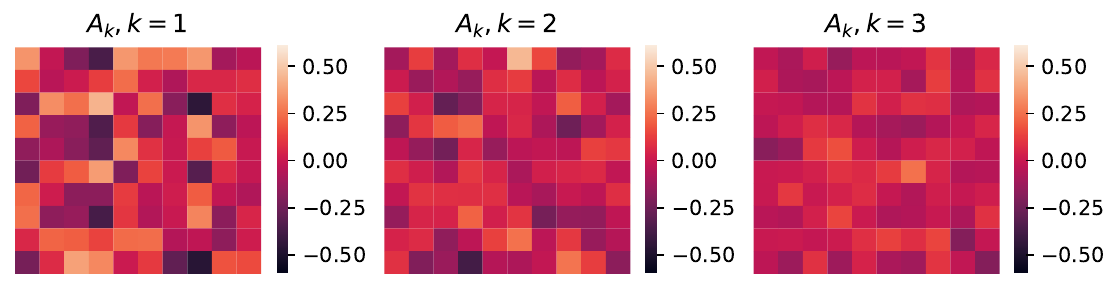}
    \caption{Coefficient matrices in NoTMF with seasonal differencing.}
    \end{subfigure}
    \begin{subfigure}{.7\linewidth}
    \centering
    \includegraphics[width=\textwidth]{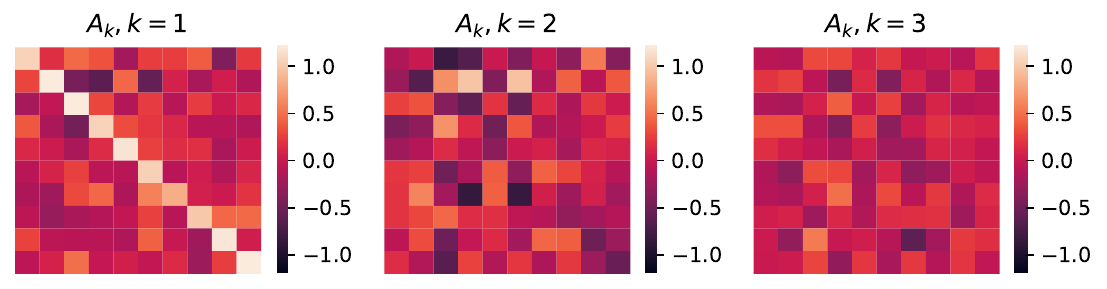}
    \caption{Coefficient matrices in NoTMF without differencing operations.}
    \end{subfigure}
\caption{Heatmap of coefficient matrices $\boldsymbol{{A}_{k}},\,\forall k\in[d]$ in NoTMF. Note that we set the rank as $R=10$ and the order as $d=3$ for both models.}
\label{coefficient_matrices}
\end{figure}

In particular, we visualize the coefficient matrix of NoTMF with the rank $R=10$ and the order $d=3$ in Figure~\ref{coefficient_matrices}. In these heatmaps, each is of size $10\times 10$ and the diagonal entries represent the auto-correlations of each time series (i.e., temporal factor). In contrast to the NoTMF model without differencing operation, it demonstrates that the coefficient matrices of NoTMF with seasonal differencing show weak correlations. This implies the importance of seasonal differencing for stationarizing the time series and eliminating the correlations.

% \begin{figure}
% \centering
% \subfigure[Coefficient matrices in NoTMF with seasonal differencing.]{
%     \centering
%     \includegraphics[width = 0.7\textwidth]{graphics/notmf_Ak.pdf}
% }
% \subfigure[Coefficient matrices in NoTMF without differencing operations.]{
%     \centering
%     \includegraphics[width = 0.7\textwidth]{graphics/tmf_Ak.pdf}
% }
% \caption{Heatmap of coefficient matrices $\{\boldsymbol{{A}_{k}}\}$ in NoTMF. Note that we set the rank as $R=10$ and the order as $d=3$ for both models.}
% \label{coefficient_matrices}
% \end{figure}

\section{Concluding Remarks}

In this study, we propose a NoTMF model, specifically designed to forecast the complex dynamics of high-dimensional, sparse, and nonstationary movement speed data collected from Uber ridesharing vehicles. Distinguishing itself from traditional TMF models, NoTMF innovatively incorporates differencing operations on latent temporal factors, enhancing its capability to model nonstationary time series data effectively. The reformulation of the temporal loss function through a sequence of matrix operators is a key innovation, enabling the application of the conjugate gradient method for solving the optimization problem. This method provides an efficient and scalable approach for approximating the closed-form least squares solution to the factor matrix from a generalized Sylvester equation, thus facilitating robust modeling and forecasting of large-scale, high-dimensional urban traffic data.
% in which the conjugate gradient method is an efficient and scalable routine for approximating the closed-form solution (i.e., least squares solution in this case) to the factor matrix from the generalized Sylvester equation. 
% Therefore, the algorithms developed in this work can support efficient modeling/forecasting on high-dimensional and large-scale movement speed data. 
Our comprehensive numerical analysis, conducted on two extensive Uber movement speed datasets, demonstrates NoTMF's superior forecasting performance compared to several baseline models.
% For the numerical analysis, the extensive experiments on two Uber movement speed datasets confirm the superior performance of NoTMF over some baseline models, and the comparison with existing TMF models shows the advantage of NoTMF for addressing the nonstationary issue before modeling time series with clear seasonality and trends.

Moreover, a comparative analysis with existing TMF models underscores NoTMF's ability to address nonstationary phenomena in time series data, particularly those exhibiting distinct seasonality and trends.
% In the modeling process, seasonal differencing can help explain seasonality and periodicity underlying movement speed data of urban road networks. 
The integration of seasonal differencing within the NoTMF framework is pivotal for elaborating the underlying seasonality and periodicity in urban road network speed data. Specifically, setting the seasonality parameter $m=1$ transforms the temporal loss function to a first-order differenced VAR process, illustrating NoTMF's flexibility in adapting to varying degrees of time series stationarity through the application of both first-order and seasonal differencing techniques.
% As a special case, if the season $m$ is set as 1, then the temporal loss in NoTMF would be referred to as the first-order differenced VAR.

Looking forward, the potential for integrating nonlinear temporal dependencies into the NoTMF model presents an exciting avenue for research. Such advancements could involve replacing the current linear latent temporal equation \eqref{obs_vs_latent_eqs} with nonlinear functions, leveraging the computational power of deep learning to capture more complex temporal dynamics in urban traffic data \citep{prince2023understanding}. This evolution of the NoTMF model would not only enhance its forecasting accuracy but also expand its applicability to a broader range of time series datasets characterized by intrinsic seasonal and trend patterns.
% To sum up, for the sake of stationarizing the time series data with trend and seasonality, one may need to introduce complicated operations such as the combination of first-order and seasonal differencing. The temporal loss in NoTMF can also be adapted for certain purposes and the resulting NoTMF can be solved by the alternating minimization mentioned in this work. Another possible direction for advancing the proposed NoTMF model would be introducing nonlinear temporal dependencies to the latent temporal factors. Accordingly, the latent equation in \eqref{obs_vs_latent_eqs} could be replaced by nonlinear functions in deep learning \citep{prince2023understanding}.

\section*{Acknowledgment}

Xinyu Chen and Chengyuan Zhang would like to thank the Institute for Data Valorisation (IVADO), Fonds de recherche du Québec -- Nature et technologies (FRQNT), and the Interuniversity Research Centre on Enterprise Networks, Logistics and Transportation (CIRRELT) for providing the PhD Excellence Scholarship to support this study.

\bibliography{references}

\end{document}